\documentclass[10pt,twocolumn,letterpaper]{article}

\PassOptionsToPackage{table}{xcolor}
\usepackage[pagenumbers]{wacv} 
\usepackage{tikz}
\usepackage{balance}

\newcommand{\ivfilter}[1]{\textsc{image adaptation#1}}
\newcommand{\cdefault}[1]{\emph{original#1}}
\newcommand{\cgray}[1]{\emph{grayscale#1}}
\newcommand{\cone}[1]{\emph{alpha-1#1}}
\newcommand{\ctwo}[1]{\emph{alpha-2#1}}
\newcommand{\cthree}[1]{\emph{alpha-3#1}}

\newcommand{\anova}[6]{{\small [$F(#1,#2)$\,$=$\,$#3$, $p$\,$#4$\,$#5$, $\eta_{p}^{2}$\,$=$\,$#6$]}}
\newcommand{\pvald}[3]{{\small ($p\,#1\,#2$, $d_z = #3$)}} 
\newcommand{\pvall}[2]{{\small $p\,#1\,#2$}} 

\definecolor{wacvblue}{rgb}{0.21,0.49,0.74}
\usepackage[pagebackref,breaklinks,colorlinks,allcolors=wacvblue]{hyperref}

\def\wacvPaperID{1168} 
\def\confName{WACV}
\def\confYear{2027}

\title{Learned Parametric Emotion Editing: Real-Time Affective Filtering for On-Device Social Media Video
}

\author{Musa Rochi$^{*}$, Marcel Schubert$^{*}$, Christoph Gebhardt\\
Eastern Switzerland University of Applied Sciences (OST)\\
{\tt\small musa.rochi@ost.ch, marcel.schubert@ost.ch, christoph.gebhardt@ost.ch}
}

\begin{document}

\maketitle

\renewcommand{\thefootnote}{\fnsymbol{footnote}}
\footnotetext[1]{Both authors contributed equally.}
\renewcommand{\thefootnote}{\arabic{footnote}}

\begin{abstract}
Problematic internet use affects a growing share of the population, yet common interventions, e.g., time limits, blocking, forced breaks, are coercive and easily circumvented. 
We explore a less restrictive alternative: adapting the emotional intensity of visual content. 
Prior work has shown that optimization can steer an image's affective content, but its per-image optimization cost makes it impractical for real-time deployment. 
We instead learn a model that predicts this transformation in a single forward pass: a MobileNetV4 backbone with FiLM-based emotion conditioning outputs parameters for differentiable global transformations.
This replaces prior iterative optimization ($\sim$80\,s per image) with a single 3.7\,ms forward pass.
In a user study ($N = 54$), the model reduced viewer-reported arousal relative to unedited images, comparably to the grayscale well-being filter, while being rated higher in perceived quality.
We integrate the model into an Android app that adapts Instagram video in real time, sustaining 60 fps on a Samsung Galaxy S23.
\end{abstract}
    
\section{Introduction}
\label{sec:introduction}

Problematic internet use (PIU) affects a growing share of the population.
PIU rates among adults are estimated at 10.1\% in the United Kingdom and 10.4\% in the United States~\cite{lopez2023problematic}, and are even higher among adolescents (e.g., 31\% in Canada \cite{lavoie2023relationship}).
In response, common interventions (e.g., time limits, content blocking, forced breaks) directly restrict access, reducing user autonomy and leading to frequent circumvention~\cite{LyngsHackMyself2020, lyngs2022goldilocks}.

An alternative direction, motivated by the established link between affect and online engagement~\cite{berger2012makes,schreiner2021impact}, is to adapt the emotional intensity of content itself rather than restrict access to it \cite{Gebhardt2025Generative}. 
Prior work has shown that optimization- and learning-based approaches can steer an image's affective content along interpretable dimensions such as valence and arousal~\cite{Gebhardt2025Generative,xia2025muse,mood2026}.
However, large generative backbones and iterative optimization procedures make them impractical for real-time, in-the-wild deployment such as filtering a live video feed.

\begin{figure}
  \centering
  \includegraphics[width=\columnwidth]{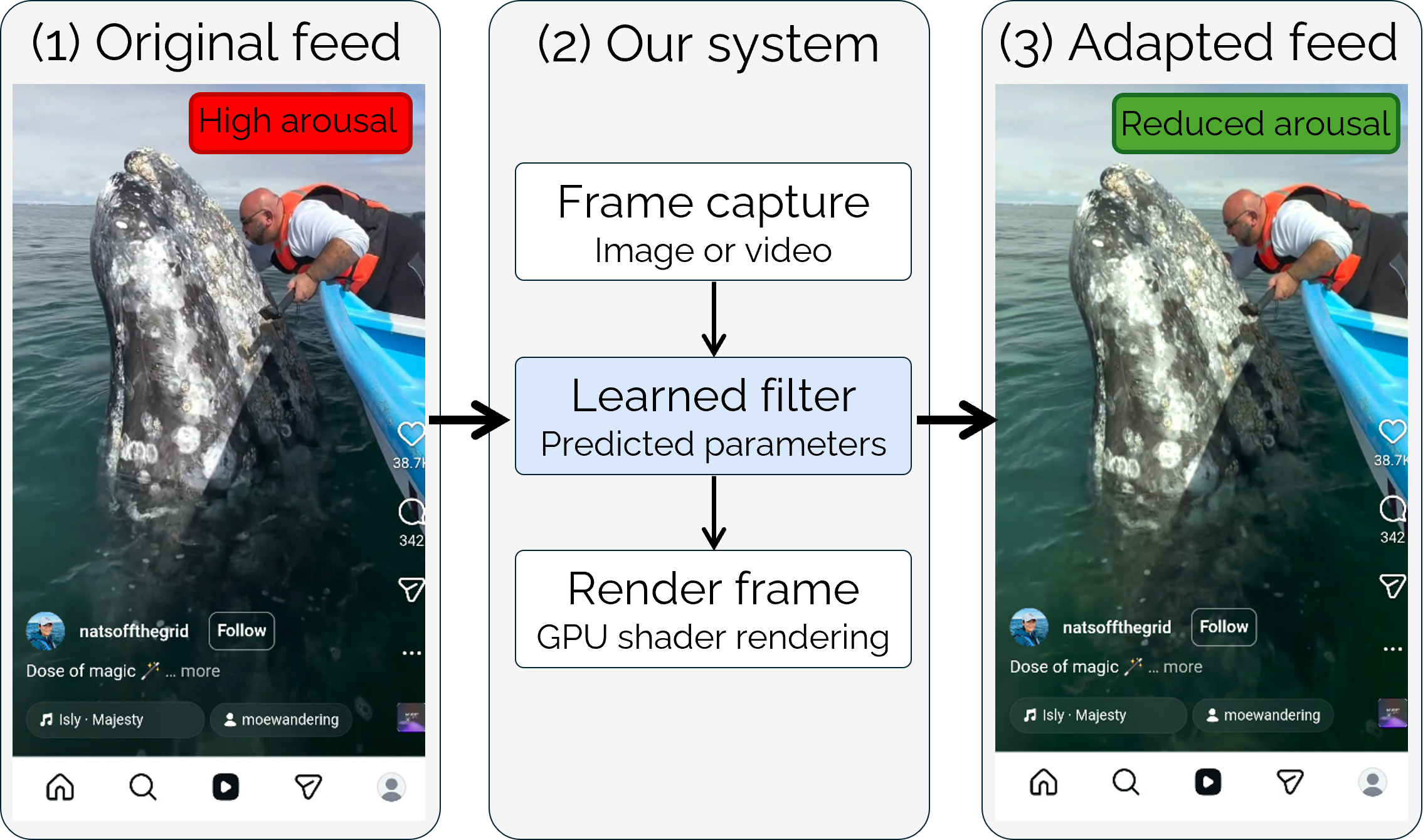}
  \caption{\textbf{Overview of our approach}: An original social media feed frame (1) is captured and processed by our learned filter (2), which predicts parameters for a sequence of differentiable global transformations in a single forward pass. The adapted frame is rendered back into the live feed (3), reducing viewer-reported arousal while preserving perceived quality.}
  \label{fig:teaser}
  \vspace{-1em}
\end{figure}

We address this gap by learning a lightweight model that predicts affective image edits in a single forward pass, replacing iterative optimization with direct parameter prediction. 
A MobileNetV4 backbone with FiLM-based emotion conditioning outputs parameters for a sequence of differentiable global transformations (e.g., saturation, contrast, tone), trained with a relative arousal-conditioning objective and CLIP-based content preservation. 
We integrate this model into a functional Android application that intercepts and adapts Instagram video in real time via WebView injection and GPU shader rendering, decoupling inference from playback to hide per-frame latency without visible artifacts.

We validate the approach in two complementary ways: benchmarks confirming the model's control over predicted arousal as well as the system's real-time performance (\cref{sec:evaluation}), and a controlled user study in which participants viewed and rated edited images (\cref{sec:user-study}). The model significantly reduced viewer-reported arousal relative to unedited images, comparably to grayscale filtering, and was rated significantly higher in perceived quality than grayscale, a filter commonly used in existing digital well-being tools.

In summary, our contributions are:

\begin{enumerate}
    \item A learned, conditionally-controllable model for affective image editing that predicts global transformation parameters in a single forward pass, replacing $\sim$80\,s of per-image iterative optimization with a 3.7\,ms forward pass ($\sim$22{,}000$\times$ faster per edit).
    
    \item A functional, real-time Android application that applies this model to live video within a social media platform (Instagram), sustaining 60 fps on a Samsung Galaxy S23 through asynchronous scheduling and temporal smoothing, at an amortized $\sim$1.27\,ms per rendered frame.
    
    \item A user study showing that the model's edits reduce perceived arousal comparably to the grayscale well-being filter, at a significantly lower perceptual-quality cost.
\end{enumerate}

Code and pretrained models will be made publicly available upon publication.

\section{Related Work}
\label{sec:related-work}
We review two lines of prior work relevant to our approach: digital self-control tools and computational methods for adapting the emotional content of images. We organize the latter into three categories: standard image transformations, GAN-based methods, and diffusion-based methods.

\subsection{Digital Self-Control Tools}
Digital self-control tools (DSCTs) help users limit habitual digital media use.
Common strategies include hiding distracting content \cite{JDev2019}, limiting functionality \cite{LessPhone2019}, gamification \cite{Forest2018}, and time-spent visualizations \cite{Apple2023}, alongside design patterns such as goal reminders \cite{ko2015nugu}, and gesture-inhibiting interfaces \cite{lu2024interactout}. 
However, such mechanisms mostly rely on simple restriction heuristics (e.g., time limits) that trigger psychological reactance
\cite{brehm2013psychological, lukoff2022designing} and are frequently circumvented \cite{LyngsHackMyself2020}, motivating interventions that regulate behavior without feeling coercive \cite{lyngs2022goldilocks}.

Closest to our work, Gebhardt et al.~\cite{Gebhardt2025Generative} adapt the emotional properties of images toward neutral valence and low arousal and show this reduces both viewers' arousal and time spent on a social media feed. However, their approach relies on iterative per-image optimization and can alter image semantics (e.g., appearance, age, skin color), raising ethical concerns. 
In contrast, our method predicts parameters for a set of low-level global transformations: since any adjustment is applied uniformly across the frame, it cannot selectively alter subject-specific attributes such as skin tone independent of the surrounding scene, preserving semantic content by construction. 

\subsection{Image Transformations}
Non-photorealistic rendering has been used to alter emotional responses by producing stylized images: such algorithms shift intense emotions toward a more neutral state, often through the introduction of confusion or loss of detail~\cite{mould2012emotional,besanccon2018reducing}. These methods do not target a specific emotional direction, unlike our approach.

Later work introduces targeted control via a reference image: early methods identify an image's emotion distribution and apply standard transformations to align it with a reference image's emotional state~\cite{ali2017automatic, peng2015mixed}, while An et al.~\cite{an2021global} instead optimize latent vectors to align an image with a reference selected via emotional vocabulary. 
Since these approaches require a reference image to define the target emotion, they are incompatible with our use case, where suitable reference images cannot be assumed.

\subsection{GAN-based Methods}
Various works use GANs to adapt image affect. Some align source images' emotional distributions with a target domain to enable cross-dataset emotion classification \cite{zhao2019cycleemotiongan, zhao2018emotiongan}. Building on content-style disentanglement \cite{huang2018multimodal,lee2018diverse}, other methods transfer the style of an emotional reference image to an input, either globally \cite{zhu2022emotional} or at the object level using semantic segmentation and multiple references \cite{chen2020image}.

Moving away from reference-based guidance, others use regressors to generate images from specified sentiments: one GAN architecture generates emotional landscape images from noise guided by valence-arousal values \cite{park2020emotional}, while another modifies a pre-trained GAN's latent noise vector to shift an image's valence, using a regressor to verify the change \cite{goetschalckx2019ganalyze}. However, both generate affective images from noise rather than adapting existing images.

Weng et al.~\cite{weng2023affective} jointly encode an image and emotional text description via a transformer, decoding with a CNN to align emotional characteristics with the text while preserving semantics.
All of the above process full-resolution images through deep convolutional architectures, incurring substantial inference cost that makes real-time, on-device deployment impractical without further compression.

\begin{figure*}[t]
    \centering
    \includegraphics[width=0.9\linewidth]{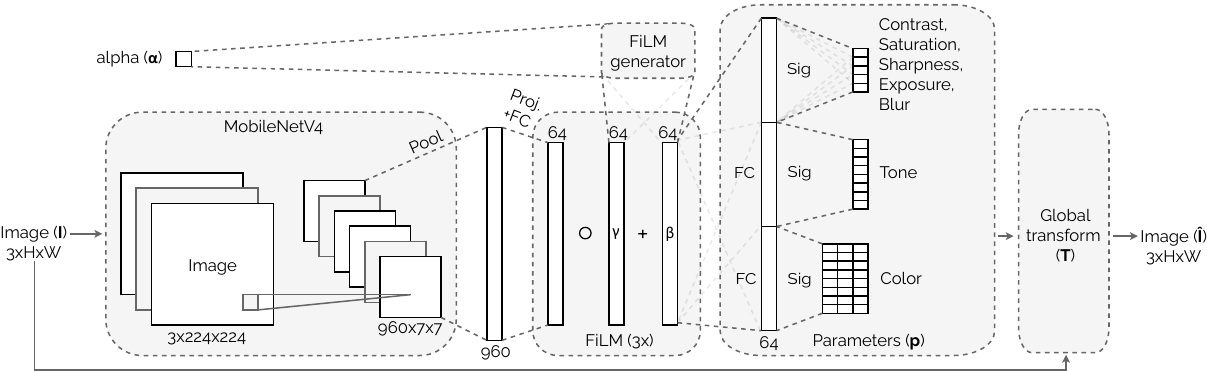}
    \caption{\textbf{Conditional parametric generator.} The conditioning signal $\alpha$ and the input image $I$ are encoded separately and fused via three FiLM blocks before three prediction heads regress the parameters $p$ of the global transformation $T$.}
    \label{fig:architecture}
\end{figure*}

\subsection{Diffusion-based Methods}
\label{sec:rw-diffusion}
Recently, diffusion models (DMs) have surpassed GANs in image synthesis quality \cite{ho2020denoising,rombach2022high}, prompting research into using DMs to modify images' emotional properties. 
Several approaches fine-tune Instruct-Pix2Pix \cite{brooks2023instructpix2pix} on emotion-annotated data to adapt an image to a target emotion while preserving scene structure, guided by a discrete emotion label \cite{yang2025emoedit,lin2025make}. 
A multi-agent framework extends this to generate multiple visually distinct yet emotionally consistent edits from the same input and target emotion \cite{mao2025emoagent}.


More recent work conditions diffusion models on continuous valence-arousal values. 
EmotiCrafter fuses valence-arousal values into textual features via an emotion-embedding network to align generated images with the prompt's emotional intent~\cite{dang2025emoticrafter}. 
EmoFeedback\textsuperscript{2} instead uses reinforcement tuning, where a vision-language model scores generated images' valence-arousal values to provide emotion-based rewards~\cite{jia2025emofeedback2}. 
MUSE optimizes an emotion token appended to the text embedding, supervised by an off-the-shelf emotion classifier applied to the denoised image estimate at each timestep~\cite{xia2025muse}. 
MooD retrieves a valence-arousal-matched reference image and use it to guide the diffusion process toward the target emotion~\cite{mood2026}.
Like all diffusion-based approaches, these methods require multiple denoising steps per image, compounding inference cost.

In summary, to the best of our knowledge, no prior method adapts the emotional content of images in real time, on-device, based on continuous arousal conditioning.
\section{Method}
\label{sec:method}
Emotions are commonly conceptualized using the Circumplex Model of Affect (CMA)~\cite{russell_circumplex_1980}, which characterizes them along two dimensions: valence, the degree of pleasantness, and arousal, the level of intensity. 
Building on evidence that reducing arousal and shifting valence toward neutral can decrease online engagement~\cite{Gebhardt2025Generative}, our approach adjusts visual content along these dimensions with the aim of reducing the time users spend online.
We focus on arousal as it is associated with a single-direction activation response, whereas valence tends to correlate with increased engagement at \emph{both} polarizing ends \cite{yu2014we,Brady2017EmotionMoralizedContent}, making a directional editing objective for it less well-posed.
%

\subsection{Problem Formulation}
\label{sec:method:problem}

Let $R$ denote a frozen affect regressor that predicts a valence--arousal pair $R(I) = [R_v(I),\, R_a(I)]^\top$ from an image $I$. As our control signal targets arousal only, we define, given a desired shift $\alpha$, the target arousal as $a' = \operatorname{clip}(R_a(I) + \alpha, 0, 1)$ and seek an edited image $\hat{I}$ such that $R_a(\hat{I})$ approximates $a'$, while (i) preserving the semantic content of $I$, (ii) remaining perceptually close to $I$, and (iii) admitting evaluation under a real-time latency budget on consumer edge hardware (\cref{sec:system}).

Rather than synthesizing $\hat{I}$ directly, we cast editing as \emph{parameter prediction}: a generator $f_\theta$ predicts the parameters $p$ of a fixed, differentiable global transformation $T$, so that $\hat{I} = T(I, p)$ realizes the desired arousal shift. 
%
%
This restricts the hypothesis space to global, low-level transformations: $T$ applies exposure, saturation, an 8-step tone curve, an 8-step per-channel color curve, contrast, sharpening, and Gaussian blur, in this order (parameterization in \cref{apx:transform}, Suppl. Mat.).
The deliberate trade-off forfeits the high-level editing capacity of GAN- or DM-based approaches in exchange for a model that is orders of magnitude cheaper to evaluate per frame (cf.~\cite{mejjati2020look}).
This choice further extends naturally to video: since $T$ is a cheap, closed-form transformation, predicted parameters can be reused across several subsequent frames without re-invoking $f_\theta$, allowing inference to run at a fraction of the video frame rate (\cref{sec:system:pipeline}).

\subsection{Conditional Parametric Generator}
\label{sec:method:architecture}

\cref{fig:architecture} shows the generator. It consists of three components: (1) a conditioning encoder that embeds the scalar $\alpha$ into a latent vector; (2) an image encoder that extracts visual features; and (3) three fully-connected prediction heads, applied to FiLM-modulated trunk representations, that regress the raw parameters.
Raw head outputs are passed through a sigmoid-based mapping that rescales them into each parameter's valid range $[m, M]$, centered such that a zero output maps to the parameter's identity value.
Internally, the generator uses leaky ReLU activations throughout.

Only the backbone operates on a fixed resolution: input images are resized to $224\times224$ before feature extraction, matching the MobileNetV4 pretraining resolution. The transform $T$, in contrast, is applied to the image at its native resolution, allowing the method to operate on arbitrary input sizes end-to-end.

For the image encoder, we depart from the backbone used in prior parametric-editing work~\cite{mejjati2020look} ($\sim$6.5M parameters) and instead use MobileNetV4~\cite{qin2024mobilenetv4}, a backbone optimized for the accuracy--latency trade-off on edge devices ($\sim$3.8M parameters, 73.8\% ImageNet top-1). 
This substitution is the main architectural driver of real-time deployment: it reduces capacity by 42\% while retaining sufficient capacity for the parameter-regression task. We initialize the encoder from ImageNet-pretrained weights~\cite{timm2024mobilenetv4}.

Naively concatenating the conditioning embedding with the image features is prone to mode collapse (e.g.\ a uniform green tint under negative $\alpha$). To mitigate this, we condition the trunk via Feature-wise Linear Modulation (FiLM)~\cite{perez2018film,dumoulin2018feature}, applying an elementwise affine transformation to the image features rather than concatenating the conditioning signal directly: $\mathrm{FiLM}(x) = \gamma(z) \odot x + \beta(z),$ where $x$ is the image feature map and $z$ the conditioning embedding. 

The backbone's average-pooled 960-dimension feature vector is projected to a 64-d shared trunk representation, which is modulated by three sequential FiLM residual blocks (see \cref{fig:architecture}). Within the FiLM generator, the conditioning value $\alpha$ is first encoded by a two-layer dense network ($1\!\to\!32\!\to\!32$); within each block, this embedding is further transformed by a dedicated modulation network ($32\!\to\!64\!\to\!128$) and split into scale and shift vectors $\gamma, \beta \in \mathbb{R}^{64}$. 
FiLM enforces a clearer separation between image content and control signal than concatenation, and extends naturally to vector-valued conditioning (e.g., joint valence-arousal conditioning) unlike multiplicative modulation schemes that typically require architectural changes.

\subsection{Training Objective}
\label{sec:method:objective}

The generator is trained end-to-end against the frozen regressor $R$ and a frozen CLIP image encoder~\cite{radford2021learning}, with no adversarial component. Given a sampled conditioning value $\alpha$, the loss is

\begin{align}
\mathcal{L}(\theta) = \mathbb{E}_{I,\alpha}\Big[
    &\big(R_a(T(I, f_\theta(I,\alpha))) - (R_a(I) + \alpha)\big)^2 \notag \\
    &+ \lambda_{\mathrm{clip}}\,\mathcal{L}_{\mathrm{CLIP}}(I, T(I,f_\theta(I,\alpha)))
\Big],
\label{eq:loss}
\end{align}

where the first term drives the predicted arousal of the edited image toward the target and the second is a semantic-preservation term,

\begin{equation}
\mathcal{L}_{\mathrm{CLIP}}(I, \hat{I}) = 1 - \frac{c \cdot \hat{c}}{\lVert c \rVert \lVert \hat{c} \rVert}, \quad c = \phi(I),\ \hat{c} = \phi(\hat{I}).
\label{eq:clip}
\end{equation}

Only $\theta$ is optimized. Gradients traverse the frozen regressor $R$, the frozen CLIP encoder $\phi$, and the differentiable transform $T$ to reach $f_\theta$; $T$ has no learnable parameters.

\subsection{Training Setup}
\label{sec:method:training}

Models are trained on MS-COCO~\cite{lin2014microsoft} for up to 100 epochs ($\sim$10M images seen), with $\alpha \sim \mathcal{U} (-0.5, 0.5)$ sampled per example and target scores clipped to $[0,1]$. 
The affect regressor $R$ is a ResNet50 fine-tuned on a custom dataset assembled from ten existing affective image databases \cite{Gebhardt2025Generative}; it is used frozen and unmodified throughout this work.
Validation loss decreased substantially over the first 25 epochs ($\sim$2.5M images seen, $\sim$3.8 GPU-hours on 8$\times$Tesla P100), capturing most of the total improvement, training then improved only marginally (3.0\%) before plateauing by epoch 60, fluctuating within 0.3\% of its mean for the remaining 40 epochs.
The full training configuration, procedure, and gradient flow, are illustrated in \cref{apx:training-figure,apx:va-dataset}.

\section{System}
\label{sec:system}

We deploy the model as a proof of concept on live short-form video inside a mobile social media application, running entirely on-device (constraints detailed in \cref{sec:system:decisions}).

\subsection{Architecture Overview}
\label{sec:system:architecture}
We implement the system as a single-container Android application: a WebView acts as a transparent proxy for the target application (e.g. Instagram), into which an Overlay System is injected via JavaScript. The Overlay System observes the DOM for video elements and delegates model inference to an ML Runtime component, which runs an ONNX representation of our model via WebAssembly.
The transformed frames are then rendered back onto the page through an overlay canvas, using GLSL fragment shaders executed via WebGL.
To avoid the additional latency of cross-process or network round-trips, we deliberately avoid a backend or hybrid native/web split.

\subsection{Real-Time Pipeline}
\label{sec:system:pipeline}

\begin{figure}[t]
    \centering
    \includegraphics[width=\linewidth]{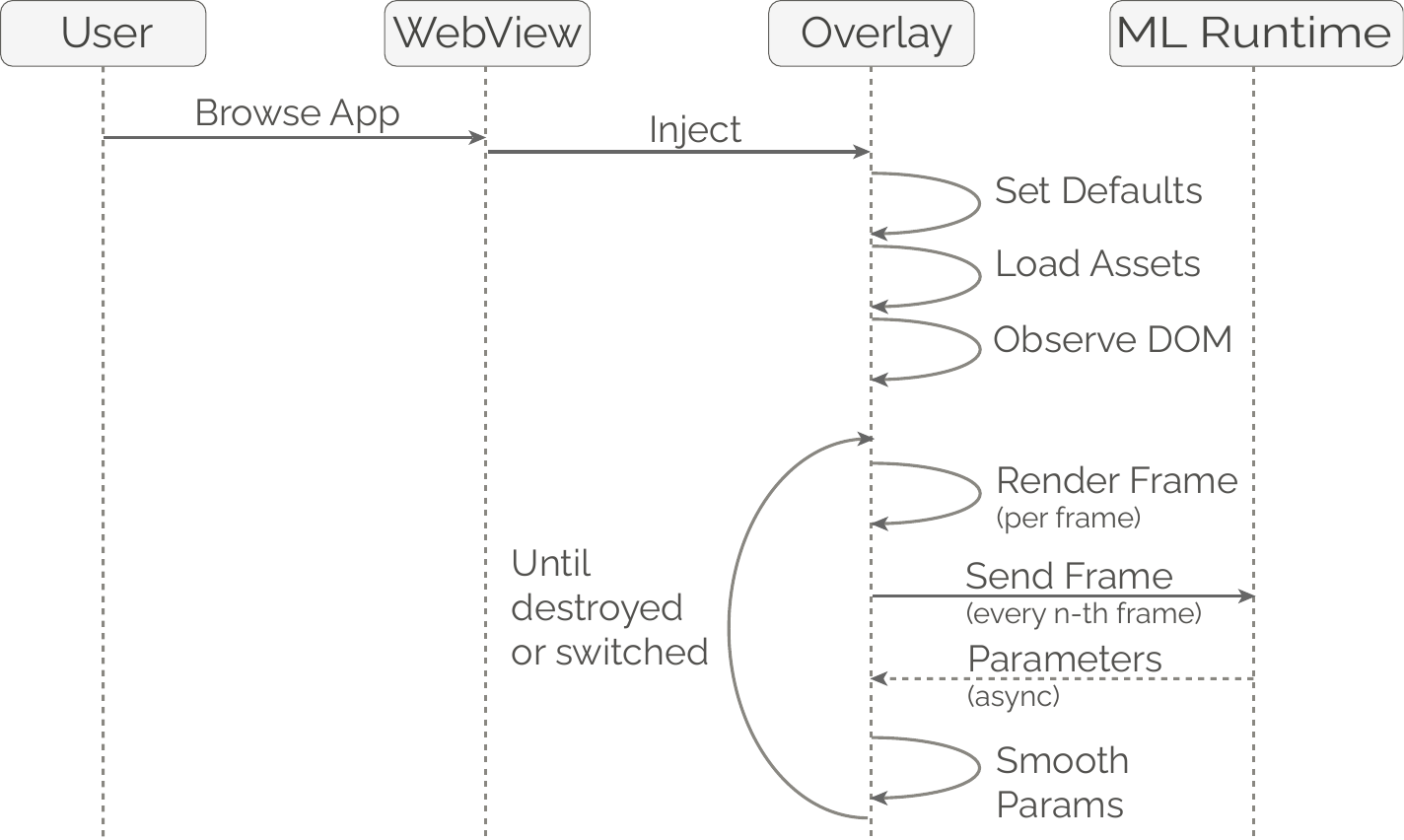}
    \caption{\textbf{End-to-end pipeline.} \texttt{Render Frame} executes on every animation frame; inference (\texttt{Send Frame} $\to$ \texttt{Parameters}) is triggered only every $n$-th frame and runs asynchronously off the main thread. Predicted parameters are smoothed (\cref{eq:smoothing}) before being applied to subsequent frames.}
    \label{fig:pipeline}
\end{figure}

Our end-to-end pipeline is illustrated in \cref{fig:pipeline}. At any time, only the \emph{active} video, the DOM video element visible in the viewport, is processed; when it changes, the previous pipeline is torn down and a new one attached.

For the active video, the renderer is invoked via \texttt{requestVideoFrameCallback}, which fires when a new video frame is presented for compositing, rather than on the display's refresh cycle~\cite{mdn_requestvideoframecallback}.
On each call, it applies the transform $T$ with the current parameter estimate $p$.

Model inference, by contrast, is triggered only once every $n{=}60$ frames and executed asynchronously on a Web Worker, off the main UI thread: the current frame is downsampled to the backbone's input resolution, copied to CPU memory, and passed to the ONNX runtime. This amortizes the cost of inference across rendered frames, exploiting the fact that short-form video content tends to be visually stable over sub-second intervals; $n$ is a tunable, device-dependent trade-off between responsiveness and inference cost.

Since parameters are only updated once every $n$ frames, applying them directly on receipt produces visible discontinuities. We therefore temporally smooth each predicted parameter independently, treating it as a 1-D signal filtered via a discrete-time spring--damper update: at every rendered frame, the parameter's velocity is accelerated toward the newest target, damped, and integrated,

\begin{equation}
v_t = d\,(v_{t-1} + k\,(x_{\mathrm{target}} - x_{t-1})), \quad
x_t = x_{t-1} + v_t,
\label{eq:smoothing}
\end{equation}

with stiffness $k$ and damping $d$ (empirically set to $k{=}0.05$, $d{=}0.5$) controlling the trade-off between responsiveness and smoothness. This applies low-pass filtering to the parameter trajectory, suppressing frame-to-frame discontinuities while smoothly converging to each new target.

\section{Evaluation}
\label{sec:evaluation}


\subsection{Model}
\label{sec:evaluation:model}

\paragraph{Architecture ablation.} 
We compare five configurations at epoch 25, varying the FiLM trunk (present vs.\ absent) and the head output mapping (sigmoid vs.\ tanh), plus a wider sigmoid-FiLM variant (trunk width 128), against Loss, Arousal MSE, and CLIP preservation loss (\cref{tab:ablation}; full training curves in \cref{apx:ablation-curves}). 
Model evaluation was conducted on a fixed set of 5{,}000 images from the COCO~2017 test set~\cite{lin2014microsoft}.
The two sigmoid-FiLM configurations achieve the lowest Loss and Arousal MSE, with the wider variant showing no gain over the base width despite its added cost. This comes at a trade-off: CLIP preservation loss follows the reverse ordering, with the no-FiLM variants achieving the lowest values and the wider sigmoid-FiLM variant the highest, indicating that FiLM conditioning improves steerability at some cost to strict content preservation. We select sigmoid-FiLM (width 64) for all subsequent experiments.

\paragraph{Loss ablation.} 
For the sigmoid-FiLM architecture, we additionally swept $\lambda_{\mathrm{clip}}$ across five values (\cref{tab:ablation}) on the same image set (full training curves in \cref{apx:clip-ablation-curves}). 
As expected, $\lambda_{\mathrm{clip}}$ trades off the two loss terms: smaller values prioritize arousal control, while larger values prioritize CLIP preservation. We select $\lambda_{\mathrm{clip}}=0.05$, as it sits near the knee of this trade-off, substantially reducing CLIP loss relative to smaller values at only a small cost in arousal MSE.

\begin{table}[b]
    \centering
    \small
    \begin{tabular}{lccc}
        \toprule
        \textbf{Configuration} & \textbf{Loss} & \textbf{Arousal MSE} & \textbf{CLIP loss} \\
        \midrule
        FiLM Sigmoid       & \textbf{0.0405} & \textbf{0.0341} & 0.1271 \\
        FiLM Sigmoid (128) & 0.0407 & 0.0342 & 0.1311 \\
        FiLM TanH          & 0.0422 & 0.0360 & 0.1236 \\
        No FiLM (Sigmoid)  & 0.0431 & 0.0373 & \textbf{0.1160} \\
        No FiLM (TanH)     & 0.0432 & 0.0374 & 0.1162 \\
        \midrule
        $\lambda_{\mathrm{clip}}=0.0$  & \textbf{0.0362} & 0.0362 & 0.1920 \\
        $\lambda_{\mathrm{clip}}=0.01$ & 0.0373 & \textbf{0.0353} & 0.1991 \\
        $\lambda_{\mathrm{clip}}=0.05$ & 0.0428 & 0.0362 & 0.1311 \\
        $\lambda_{\mathrm{clip}}=0.1$  & 0.0506 & 0.0417 & 0.0892 \\
        $\lambda_{\mathrm{clip}}=0.5$  & 0.0612 & 0.0573 & \textbf{0.0076} \\
        \bottomrule
    \end{tabular}
    \caption{Validation metrics at epoch 25 across five architecture configurations (top) and five different $\lambda_{\mathrm{clip}}$-values for the FiLM Sigmoid architecture (bottom); the two blocks come from separate training runs. Loss $=$ Arousal MSE $+ \lambda_{\mathrm{clip}}\cdot$CLIP loss (\cref{eq:loss}); CLIP loss as defined in \cref{eq:clip}.
    }
    \label{tab:ablation}
\end{table}

\begin{figure*}[t]
    \centering
    \includegraphics[width=\linewidth]{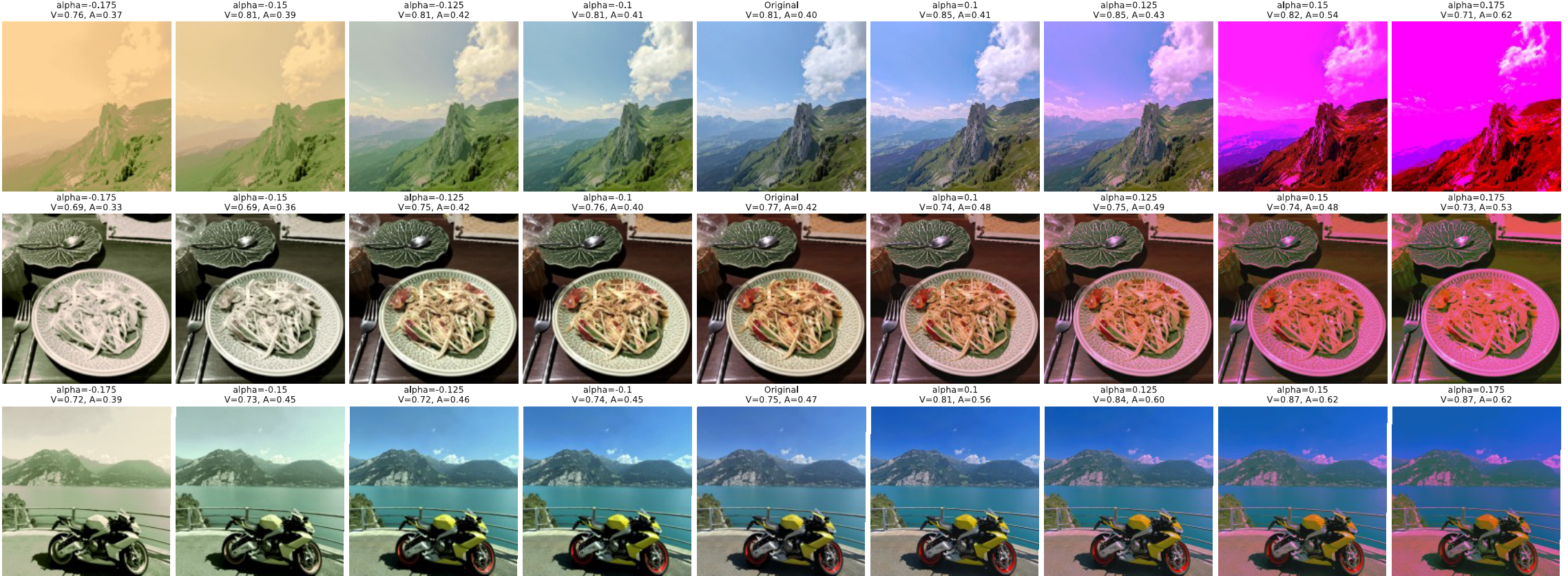}
    \caption{\textbf{Qualitative results.} Example edits for (a) decreasing $\alpha$ (vintage, desaturated shift) and (b) increasing $\alpha$ (magenta/pink shift), annotated with regressor-predicted valence (V) and arousal (A) per image. 
    }
    \label{fig:qualitative}
\end{figure*}

\paragraph{Baseline comparison.}
We compare our model against the parametric and latent style optimization baselines of Gebhardt~et~al.~\cite{Gebhardt2025Generative}, adapted to condition on the same relative arousal target $\alpha$ used by our model, rather than their original fixed valence--arousal reference. This isolates the comparison to the effect of inference versus iterative optimization, holding the conditioning mechanism fixed. 
Note that we do not benchmark against \cite{mejjati2020look}, as it was designed for a different editing task.
All methods are evaluated on a fixed, paired 500-image COCO validation subset at $\alpha \in \{-0.4, -0.2, -0.1, 0, 0.1, 0.2, 0.4\}$. 

\begin{figure}[b]
    \centering
    \begin{subfigure}[b]{0.495\linewidth}
        \centering
        \includegraphics[width=\linewidth]{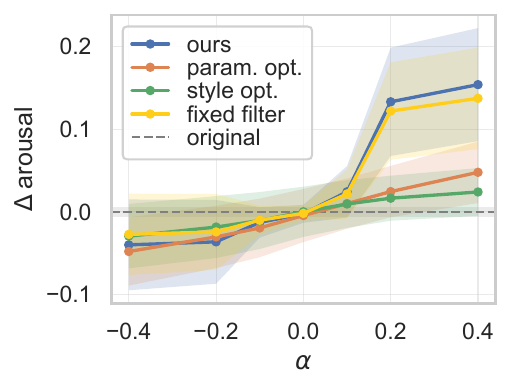}
        \caption{Control fidelity}
        \label{fig:delta-arousal}
    \end{subfigure}
    \hfill
    \begin{subfigure}[b]{0.495\linewidth}
        \centering
        \includegraphics[width=\linewidth]{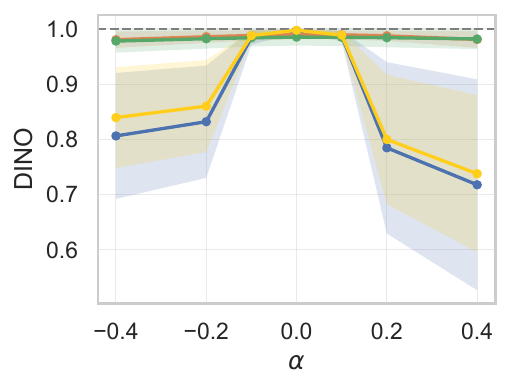}
        \caption{Content preservation}
        \label{fig:dino}
    \end{subfigure}
    \caption{\textbf{Baseline comparison.} (a) Achieved $\Delta$~arousal vs.\ requested $\alpha$; our model's response is closer to the identity line $\Delta\text{arousal}=\alpha$ than baselines. (b) Content preservation (DINO) vs.\ $\alpha$; our model achieves higher similarity near $\alpha=0$ but drops more sharply than baselines beyond $|\alpha|\approx0.1$--$0.2$.
    }
    \label{fig:baseline-comparison}
\end{figure}

\cref{fig:delta-arousal} illustrates achieved $\Delta$~arousal over $\alpha$.
As in~\cite{goetschalckx2019ganalyze}, we characterize the response with ordinary least-squares fits.
Our model's response is substantially more discernible relative to per-image noise than either baseline, with a slope roughly $2.2\times$ steeper than the parametric optimizer ($0.275$ vs.\ $0.124$; both $p<0.001$) and $3.8\times$ steeper than the style optimizer ($0.275$ vs.\ $0.073$; $p<0.001$), and a correspondingly higher coefficient of determination ($R^2=0.59$ vs.\ $0.43$ and $0.23$, respectively), indicating that both optimization baselines under-shoot the requested arousal shift.
Note that our model and the parametric optimization both outperform the latent-space optimization, suggesting that parametric, differentiable global transformations are adequate for reliable arousal control in this use case.

As image-agnostic baseline, we distill our model into a fixed filter by averaging its predicted parameters per $\alpha$ over a disjoint COCO subset. 
It outperforms both optimization baselines in slope (0.233, $R^2=0.55$, $p<0.001$), yet remains 15.3\% shallower than the full model (0.275; \cref{fig:delta-arousal}), indicating that per-image conditioning contributes steering strength beyond what a fixed transformation captures.

We report DINO (CLS-token cosine similarity~\cite{caron2021emerging}) rather than CLIP similarity to avoid overlap with the training and optimization objectives, and because DINO similarity between source and edited images is a human-aligned proxy for content preservation~\cite{basu2023editval}.
The stronger control of our approach comes at a measurable cost in preservation (\cref{fig:dino}). At $\alpha=0$, our model achieves near-perfect content preservation (DINO similarity $0.995$), exceeding both baselines' null-edit similarity ($0.985 - 0.989$).
As $|\alpha|$ increases, however, DINO similarity drops more sharply for our model than for either baseline, falling below both beyond
$|\alpha|\approx 0.1$ and reaching $0.72$--$0.81$ at $|\alpha|=0.4$, compared to the baselines' near-flat $0.98$--$0.99$ across the full range.
This is consistent with the baselines' comparatively weaker arousal response. 
Our model's DINO-based content-preservation scores ($0.72$--$0.995$) fall in a range comparable to that reported for color-editing methods evaluated with DINO~\cite{yin2024coloredit}.
These results characterize an operating boundary for our approach that we respect when generating results for our user study. 
We examine this fidelity--preservation trade-off further in \cref{apx:linear-control,apx:preservation}.

\paragraph{Qualitative results.} \cref{fig:qualitative} shows edits across a range of $\alpha$ for three example images from COCO. Negative $\alpha$ produces a desaturated, muted palette reminiscent of a vintage photograph, consistent with lower saturation and colorfulness correlating with reduced arousal~\cite{bekhtereva2017bringing, redies2020global}. 
Positive $\alpha$ instead shifts images toward a saturated pink cast, most visible in sky and background regions, consistent with red hues and increased chroma both correlating with heightened arousal~\cite{elliot2015color, elliot2019historically}.
We verify this further quantitatively in \cref{apx:attributes}.
At the extremes of the operating range, edits visibly overshoot into oversaturated pink casts (\cref{fig:qualitative}, right side), consistent with the DINO drop in \cref{fig:dino}.

\begin{figure*}[t]
    \centering
    \includegraphics[width=0.7\linewidth]{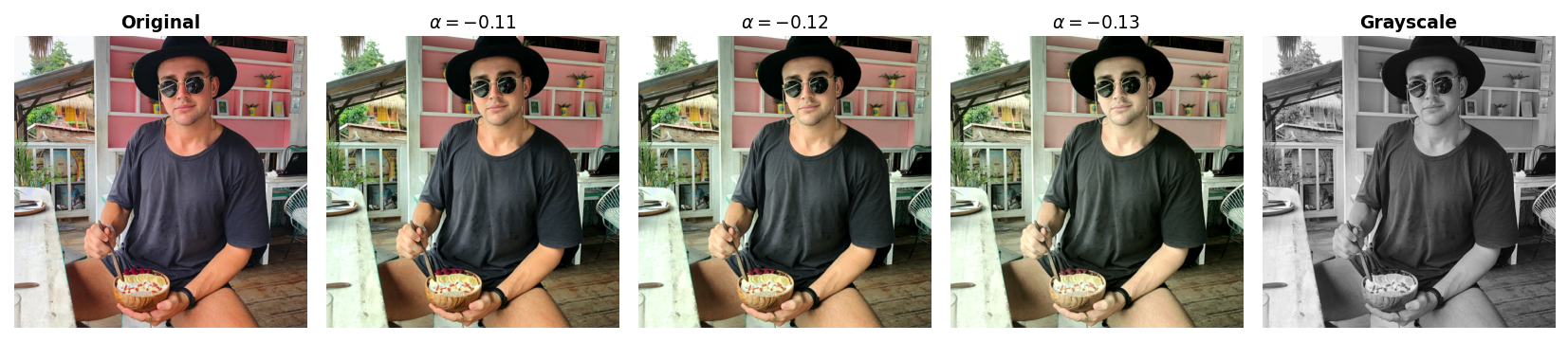}
    \caption{\textbf{Example stimulus of user study.} Stimulus across the five study conditions: original images, our approach at $\alpha = -0.11$, $\alpha = -0.12$, and $\alpha = -0.13$, and the grayscale baseline. 
    }
    \label{fig:stimulus}
\end{figure*}

\subsection{System}
\label{sec:evaluation:system}

\paragraph{Model latency.} On a single Tesla P100 GPU at $480{\times}480$, our generator achieves a median per-frame latency of 3.71\,ms (p95 4.34\,ms; $n=5{,}000$, batch size 8), while a full edit with the parametric optimization of~\cite{Gebhardt2025Generative} takes a median of 80.92\,s (p95 83.01\,s; $n=50$ images, 5 repetitions, 300 steps, single-image). This is a $\sim$22{,}000$\times$ speedup per edit (\cref{apx:latency}). 
The authors report 17.81\,s per image on an NVIDIA H200 at $1024{\times}1024$; the difference is consistent with the substantially higher throughput of their GPU. 
We do not benchmark their style optimization, which they report as $\sim$4$\times$ slower than the parametric variant, making our comparison conservative.

\paragraph{On-device latency.} 
On the target device (Samsung Galaxy S23), inference is slower: median total latency of 46.00\,ms (p95 77.80\,ms), decomposed into canvas copy (0.00\,ms, p95 0.10\,ms), pixel extraction (7.90\,ms, p95 17.00\,ms), and inference (37.90\,ms, p95 61.60\,ms; \cref{apx:latency}). 
A synchronous 16.67\,ms (60\,fps) budget is infeasible here, motivating our decoupled design: the pipeline runs asynchronously on a Web Worker only every $n{=}60$ frames, while the render loop reapplies the last smoothed parameters via the shader.
We could not instrument the shader in isolation on the target device and hence assume its cost to be ${\sim}0.5\,$ms, as measured under WebGL2 on desktop hardware (Sec.~B.1).
On that assumption, the pipeline amortizes to $(46.00 + 60\times0.5)/60 \approx 1.27\,$ms per frame, well within budget. 
Even at ten times the assumed shader cost, the render path stays under a third of the frame budget.

\paragraph{On-device energy cost.}
Across five matched sessions (same video content, filter on vs.\ off, alternated order) on a Samsung Galaxy S23, the filter increased app-attributed energy draw by a mean of $31.9 \pm 10.7$\,mAh per 30-minute session, predominantly attributable to CPU cost from model inference ($25.1 \pm 13.8$\,mAh), with no consistent change in screen- or network-attributed draw (\cref{apx:energy}). This corresponds to ca. 0.8\% of device battery capacity per session.
\section{User Study}
\label{sec:user-study}
Our technical evaluation shows the model shifts arousal as measured by the regressor. In this study, we test whether this translates to human perception, i.e., whether our adaptations reduce \emph{perceived} arousal while maintaining \emph{perceived} image quality.
As a baseline, we compare against grayscale conversion, a digital well-being intervention already deployed in mobile operating systems (e.g., Android's productivity mode, Apple's color filters). 
Full study details and statistics are provided in \cref{apx:user-study}.

\paragraph{Design.}
The study replicates the task, metrics, and hypotheses of a prior study~\cite{Gebhardt2025Generative}, differing in stimuli and generated image adaptations. It uses a within-subject design with one factor (\ivfilter{}) and five levels: \cdefault{}, \cgray{}, and our approach at three adaptation strengths, \cone{} ($\alpha = -0.11$), \ctwo{} ($\alpha = -0.12$), and \cthree{} ($\alpha = -0.13$).
We selected these $\alpha$-values based on the identified operating range in the technical evaluation, balancing arousal reduction against fidelity (\cref{fig:stimulus}). 

To ensure external validity, stimuli were 12 images drawn from the Instagram Influencer Dataset~\cite{kim2020multimodal}, shown in all conditions with Latin-square counterbalancing of condition order. 
All study stimuli were selected prior to inspecting model outputs, precluding cherry-picking.
After viewing each image, participants reported valence and arousal on a validated self-assessment scale~\cite{bradley1994measuring} and rated perceived image quality (adapted from~\cite{mould2012emotional}). 

According to an a priori power analysis, we recruited 57 participants (24 female, 33 male; ages 22--71); after pre-defined exclusions, the data of 54 people were analyzed. Significance is unaffected by exclusions.

\begin{table}[b]
  \centering
  \small
  \setlength{\tabcolsep}{4pt}
  \begin{tabular}{@{}lccc@{}}
    \toprule
    Condition   & Arousal & Quality & Valence \\
    \midrule
    \cdefault{} & 4.55 (1.98) & 3.97 (0.68) & 6.34 (1.33) \\
    \rowcolor{gray!15}
    \cone{}     & 4.36 (1.94)$^{*}$ & 3.79 (0.76)$^{*\dagger}$ & 6.18 (1.43)$^{*}$ \\
    \ctwo{}     & 4.34 (1.98)$^{*}$ & 3.71 (0.78)$^{**}$ & 6.12 (1.52)$^{*}$ \\
    \cthree{}   & 4.28 (2.10)$^{*}$ & 3.40 (0.92)$^{***\dagger}$ & 6.01 (1.56)$^{***}$ \\
    \cgray{}    & 4.19 (1.97)$^{**}$ & 3.64 (0.74)$^{***}$ & 6.05 (1.28)$^{**}$ \\
    \bottomrule
  \end{tabular}
      \caption{\textbf{Rating statistics.} Mean ratings (SD in parentheses) for arousal, perceived image quality, and valence per condition. Asterisks mark differences from \cdefault{} and daggers from \cgray{} ($^{*}p<.05$, $^{**}p<.01$, $^{***}p<.001$, $^{\dagger}p<.05$). 
      The shaded row marks our recommended operating point.
      }
  \label{tab:msd}
\end{table}

\paragraph{Results.}
\label{sec:user-study:results}
We analyzed ratings with sphericity-corrected repeated-measures ANOVAs, followed by Holm-corrected Wilcoxon signed-rank comparisons against \cdefault{} and \cgray{}. \cref{tab:msd} reports mean ratings.

\ivfilter{} affected arousal \anova{3.07}{162.78}{2.78}{=}{.042}{.05}. All conditions received significantly lower arousal ratings than \cdefault{}: \cgray{} \pvald{=}{.002}{-.42}, \cone{} \pvald{=}{.021}{-.30}, \ctwo{} \pvald{=}{.021}{-.25}, and \cthree{} \pvald{=}{.021}{-.28}. No significant differences were found between our conditions and \cgray{}.

\ivfilter{} also affected perceived quality \anova{2.78}{147.10}{18.15}{<}{.001}{.26}. While all conditions were rated below \cdefault{}, \cone{} was rated significantly \emph{higher} in quality than \cgray{} \pvald{=}{.015}{.41}, whereas \cthree{} was rated lower \pvald{=}{.021}{-.39}.

Finally, \ivfilter{} affected valence \anova{2.73}{144.62}{4.35}{=}{.007}{.08}, with all conditions rated below \cdefault{}.
This shift can be explained with the valence--arousal coupling in affective image datasets, under which reducing arousal could affect valence (\cref{apx:va-dataset}). 
Complete statistics are given in \cref{apx:user-study:results}.

\section{Discussion \& Limitations}
\label{sec:discussion}

\paragraph{User study.}
With the right parametrization ($\alpha = -0.11$), our approach reduces arousal compared to the original images and is not perceived different from the deployed grayscale baseline, while incurring a lower perceptual-quality cost. 
These effects should be read with their boundaries in mind: like grayscale, all adapted conditions still reduced perceived quality relative to the unmodified originals, arousal effect sizes were small ($|d| = 0.25$--$0.30$), and our study measured self-reported affect after single exposures in a controlled setting with 12 stimuli. 
Whether these perceptual shifts translate into behavioral outcomes, such as reduced engagement, therefore remains the central open question. Longitudinal, in-the-wild studies are the natural next step and the real-time capability of our system provides, for the first time, the infrastructure to run them.
 
\paragraph{Control fidelity.}
Our model shifts predicted arousal in the conditioned direction, but saturates and introduces artifacts beyond $|\alpha| \approx 0.15$--$0.2$. We view this ceiling as a feature as much as a limitation for our use case, which requires subtle, identity-preserving interventions.

\paragraph{Fixed filter.} Averaging our model's predictions yields an image-agnostic filter with a 15.3\% shallower response but no inference cost, running entirely in the shader ($\sim$0.5\,ms). Whether this trade-off is perceptible is untested and determines which is the better deployment target.

\paragraph{Univariate control.}
Our model is conditioned on arousal only. 
A multivariate formulation that conditions on both dimensions of the CMA could enable controlling adaptations to achieve neutral valence, which could further help to reduce online engagement.
The FiLM-based conditioning makes this extension architecturally straightforward.
 
 
\paragraph{System limitations.}
On-device inference currently runs on the CPU via WebAssembly, reaching 37.9\,ms on a Samsung Galaxy S23; GPU execution and adaptive inference scheduling offer clear headroom. The prototype further processes frames independently without explicit temporal coherence, and applies a fixed conditioning value rather than content- or user-adaptive control.
\section{Ethical Considerations}
\label{sec:ethics}

Emotion-modulating image adaptation is a dual-use capability. We designed our system as a \emph{user-initiated} digital well-being intervention: adaptation is visibly indicated, can only be toggled by the user and the app runs entirely on-device so no third party can control it. 
Applied without the user's knowledge, the same technique could instead increase arousal to drive engagement, or manipulate perception without consent.
We treat transparency and user control as prerequisites for any deployment of this technique.

Content creators' aesthetic choices are altered by the filter, though only locally at consumption time and without changing image identity, comparable to existing well-being filters (e.g., grayscale mode). Conversely, the approach can be read as returning agency to consumers, routinely exposed to content already optimized for maximal engagement.

Our user study used informed consent and compensated participation, collected only anonymous, non-health-related ratings, and did not require institutional ethics board review under applicable regulations (\cref{apx:user-study-design}).

\section{Conclusion}
\label{sec:conclusion}
We presented a learned, parametric approach to emotion editing of images that replaces per-image iterative optimization with a single forward pass.
Conditioned on a target arousal shift, the model predicts parameters of differentiable global image transformations, trained against a pre-trained affect regressor with a CLIP-based preservation term. This design reduces per-image adaptation from roughly 80\,s of iterative optimization to a 3.7\,ms forward pass and enables, for the first time, real-time affective adaptation of live social media content on a consumer smartphone, providing in turn the infrastructure to move from perceptual evidence toward behavioral evidence. 
A user study ($N = 54$) confirms that the learned edits transfer from regressor space to human perception: they reduce perceived arousal on real Instagram imagery like the deployed grayscale baseline, at a significantly lower perceptual-quality cost. Together, these results establish learned affective filtering as a practical building block for non-coercive digital well-being interventions.

{
    \small
    \bibliographystyle{ieeenat_fullname}
    \bibliography{main}

@article{lopez2023problematic,
  title={Problematic internet use among adults: A cross-cultural study in 15 countries},
  author={Lopez-Fernandez, Olatz and Romo, Lucia and Kern, Laurence and Rousseau, Am{\'e}lie and Lelonek-Kuleta, Bernadeta and Chwaszcz, Joanna and M{\"a}nnikk{\"o}, Niko and Rumpf, Hans-J{\"u}rgen and Bischof, Anja and Kir{\'a}ly, Orsolya and others},
  journal={Journal of Clinical Medicine},
  volume={12},
  number={3},
  pages={1027},
  year={2023},
  publisher={MDPI}
}

@article{lavoie2023relationship,
  title={The relationship between problematic internet use and anxiety disorder symptoms in youth: specificity of the type of application and gender},
  author={Lavoie, Christine and Dufour, Magali and Berbiche, Djamal and Therriault, Danyka and Lane, Julie},
  journal={Computers in Human Behavior},
  volume={140},
  pages={107604},
  year={2023},
  publisher={Elsevier}
}

@article{russell_circumplex_1980,
	title = {A circumplex model of affect},
	volume = {39},
	doi = {10.1037/h0077714},
	journal = {Journal of Personality and Social Psychology},
	author = {Russell, James A.},
	year = {1980},
	pages = {1161--1178},
}

@article{schreiner2021impact,
  title={Impact of content characteristics and emotion on behavioral engagement in social media: literature review and research agenda},
  author={Schreiner, Melanie and Fischer, Thomas and Riedl, Rene},
  journal={Electronic Commerce Research},
  volume={21},
  pages={329--345},
  year={2021},
  publisher={Springer}
}

@article{berger2012makes,
  title={What makes online content viral?},
  author={Berger, Jonah and Milkman, Katherine L},
  journal={Journal of marketing research},
  volume={49},
  number={2},
  pages={192--205},
  year={2012},
  publisher={SAGE Publications Sage CA: Los Angeles, CA}
}

@article{yu2014we,
  title={We look for social, not promotion: Brand post strategy, consumer emotions, and engagement},
  author={Yu, Jusheng},
  journal={International Journal of Media \& Communication},
  volume={1},
  number={2},
  pages={28--37},
  year={2014},
  publisher={Citeseer}
}

@article{
Brady2017EmotionMoralizedContent,
author = {William J. Brady  and Julian A. Wills  and John T. Jost  and Joshua A. Tucker  and Jay J. Van Bavel },
title = {Emotion shapes the diffusion of moralized content in social networks},
journal = {Proc. of the National Academy of Sciences},
volume = {114},
number = {28},
pages = {7313-7318},
year = {2017},
doi = {10.1073/pnas.1618923114}
}

@article{bekhtereva2017bringing,
  title={Bringing color to emotion: The influence of color on attentional bias to briefly presented emotional images},
  author={Bekhtereva, Valeria and M{\"u}ller, Matthias M},
  journal={Cognitive, Affective, \& Behavioral Neuroscience},
  volume={17},
  number={5},
  pages={1028--1047},
  year={2017},
  publisher={Springer}
}

@article{redies2020global,
  title={Global image properties predict ratings of affective pictures},
  author={Redies, Christoph and Grebenkina, Maria and Mohseni, Mahdi and Kaduhm, Ali and Dobel, Christian},
  journal={Frontiers in psychology},
  volume={11},
  pages={953},
  year={2020},
  publisher={Frontiers Media SA}
}

@article{specker2018universal,
  title={The universal and automatic association between brightness and positivity},
  author={Specker, Eva and Leder, Helmut and Rosenberg, Raphael and Hegelmaier, Lisa Mira and Brinkmann, Hanna and Mikuni, Jan and Kawabata, Hideaki},
  journal={Acta Psychologica},
  volume={186},
  pages={47--53},
  year={2018},
  publisher={Elsevier}
}

@article{gao2007analysis,
  title={Analysis of cross-cultural color emotion},
  author={Gao, Xiao-Ping and Xin, John H and Sato, Tetsuya and Hansuebsai, Aran and Scalzo, Marcello and Kajiwara, Kanji and Guan, Shing-Sheng and Valldeperas, Josep and Lis, Manuel Jos{\'e} and Billger, Monica},
  journal={Color Research \& Application},
  volume={32},
  number={3},
  pages={223--229},
  year={2007},
  publisher={Wiley Online Library}
}

@article{yang2020emotion,
  title={Emotion variation from controlling contrast of visual contents through EEG-Based deep emotion recognition},
  author={Yang, Heekyung and Han, Jongdae and Min, Kyungha},
  journal={Sensors},
  volume={20},
  number={16},
  pages={4543},
  year={2020},
  publisher={MDPI}
}

@article{cano2009affective,
  title={Affective valence, stimulus attributes, and P300: color vs. black/white and normal vs. scrambled images},
  author={Cano, Maya E and Class, Quetzal A and Polich, John},
  journal={International Journal of Psychophysiology},
  volume={71},
  number={1},
  pages={17--24},
  year={2009},
  publisher={Elsevier}
}

@article{simola2015affective,
  title={Affective processing in natural scene viewing: Valence and arousal interactions in eye-fixation-related potentials},
  author={Simola, Jaana and Le Fevre, Kevin and Torniainen, Jari and Baccino, Thierry},
  journal={NeuroImage},
  volume={106},
  pages={21--33},
  year={2015},
  publisher={Elsevier}
}

@article{delplanque2007spatial,
  title={Spatial frequencies or emotional effects?: A systematic measure of spatial frequencies for IAPS pictures by a discrete wavelet analysis},
  author={Delplanque, Sylvain and N’diaye, Karim and Scherer, Klaus and Grandjean, Didier},
  journal={Journal of neuroscience methods},
  volume={165},
  number={1},
  pages={144--150},
  year={2007},
  publisher={Elsevier}
}

@article{de2010effects,
  title={Effects of picture size reduction and blurring on emotional engagement},
  author={De Cesarei, Andrea and Codispoti, Maurizio},
  journal={PloS One},
  volume={5},
  number={10},
  pages={e13399},
  year={2010},
  publisher={Public Library of Science San Francisco, USA}
}

@inproceedings{mejjati2020look,
  title={Look here! a parametric learning based approach to redirect visual attention},
  author={Mejjati, Youssef A and Gomez, Celso F and Kim, Kwang In and Shechtman, Eli and Bylinskii, Zoya},
  booktitle={European Conf. on Computer Vision},
  pages={343--361},
  year={2020},
  organization={Springer}
}

@inproceedings{lee2018diverse,
  title={Diverse image-to-image translation via disentangled representations},
  author={Lee, Hsin-Ying and Tseng, Hung-Yu and Huang, Jia-Bin and Singh, Maneesh and Yang, Ming-Hsuan},
  booktitle={Proc. of the European conf. on computer vision (ECCV)},
  pages={35--51},
  year={2018}
}

@inproceedings{huang2018multimodal,
  title={Multimodal unsupervised image-to-image translation},
  author={Huang, Xun and Liu, Ming-Yu and Belongie, Serge and Kautz, Jan},
  booktitle={Proc. of the European conf. on computer vision (ECCV)},
  pages={172--189},
  year={2018}
}

@inproceedings{rombach2022high,
  title={High-resolution image synthesis with latent diffusion models},
  author={Rombach, Robin and Blattmann, Andreas and Lorenz, Dominik and Esser, Patrick and Ommer, Bj{\"o}rn},
  booktitle={Proc. of the IEEE/CVF Conf. on Computer Vision and Pattern Recognition},
  pages={10684--10695},
  year={2022}
}

@article{ho2020denoising,
  title={Denoising diffusion probabilistic models},
  author={Ho, Jonathan and Jain, Ajay and Abbeel, Pieter},
  journal={Advances in Neural Information Processing Systems},
  volume={33},
  pages={6840--6851},
  year={2020}
}

@inproceedings{brooks2023instructpix2pix,
  title={Instructpix2pix: Learning to follow image editing instructions},
  author={Brooks, Tim and Holynski, Aleksander and Efros, Alexei A},
  booktitle={Proc. of the IEEE/CVF Conf. on Computer Vision and Pattern Recognition},
  pages={18392--18402},
  year={2023}
}

@inproceedings{besanccon2018reducing,
  title={Reducing affective responses to surgical images through color manipulation and stylization.},
  author={Besan{\c{c}}on, Lonni and Semmo, Amir and Biau, David and Frachet, Bruno and Pineau, Virginie and Sariali, El Hadi and Taouachi, Rabah and Isenberg, Tobias and Dragicevic, Pierre},
  booktitle={Expressive},
  pages={11--1},
  year={2018}
}

@article{mould2012emotional,
  title={Emotional response and visual attention to non-photorealistic images},
  author={Mould, David and Mandryk, Regan L and Li, Hua},
  journal={Computers \& Graphics},
  volume={36},
  number={6},
  pages={658--672},
  year={2012},
  publisher={Elsevier}
}

@article{zhu2022emotional,
  title={Emotional Generative Adversarial Network for Image Emotion Transfer},
  author={Zhu, Siqi and Qing, Chunmei and Chen, Canqiang and Xu, Xiangmin},
  journal={Expert Systems with Applications},
  pages={},
  year={2022},
  publisher={Elsevier}
}

@article{ali2017automatic,
  title={Automatic image transformation for inducing affect},
  author={Ali, Afsheen Rafaqat and Ali, Mohsen},
  journal={arXiv:1707.08148},
  year={2017}
}

@inproceedings{peng2015mixed,
  title={A mixed bag of emotions: Model, predict, and transfer emotion distributions},
  author={Peng, Kuan-Chuan and Chen, Tsuhan and Sadovnik, Amir and Gallagher, Andrew C},
  booktitle={Proc. of the IEEE conf. on computer vision and pattern recognition},
  pages={860--868},
  year={2015}
}

@inproceedings{an2021global,
  title={Global Image Sentiment Transfer},
  author={An, Jie and Chen, Tianlang and Zhang, Songyang and Luo, Jiebo},
  booktitle={2020 25th International Conf. on Pattern Recognition (ICPR)},
  pages={6267--6274},
  year={2021},
  organization={IEEE}
}

@inproceedings{goetschalckx2019ganalyze,
  title={Ganalyze: Toward visual definitions of cognitive image properties},
  author={Goetschalckx, Lore and Andonian, Alex and Oliva, Aude and Isola, Phillip},
  booktitle={Proc. of the ieee/cvf international conf. on computer vision},
  pages={5744--5753},
  year={2019}
}

@inproceedings{park2020emotional,
  title={Emotional landscape image generation using generative adversarial networks},
  author={Park, Chanjong and Lee, In-Kwon},
  booktitle={Proc. of the Asian Conf. on Computer Vision},
  year={2020}
}

@inproceedings{zhao2019cycleemotiongan,
  title={Cycleemotiongan: Emotional semantic consistency preserved cyclegan for adapting image emotions},
  author={Zhao, Sicheng and Lin, Chuang and Xu, Pengfei and Zhao, Sendong and Guo, Yuchen and Krishna, Ravi and Ding, Guiguang and Keutzer, Kurt},
  booktitle={Proc. of the AAAI conf. on artificial intelligence},
  volume={33},
  number={01},
  pages={2620--2627},
  year={2019}
}

@inproceedings{zhao2018emotiongan,
  title={Emotiongan: Unsupervised domain adaptation for learning discrete probability distributions of image emotions},
  author={Zhao, Sicheng and Zhao, Xin and Ding, Guiguang and Keutzer, Kurt},
  booktitle={Proc. of the 26th ACM international conf. on Multimedia},
  pages={1319--1327},
  year={2018}
}

@inproceedings{chen2020image,
  title={Image sentiment transfer},
  author={Chen, Tianlang and Xiong, Wei and Zheng, Haitian and Luo, Jiebo},
  booktitle={Proc. of the 28th ACM International Conf. on Multimedia},
  pages={4407--4415},
  year={2020}
}

@inproceedings{yang2025emoedit,
  title={Emoedit: Evoking emotions through image manipulation},
  author={Yang, Jingyuan and Feng, Jiawei and Luo, Weibin and Lischinski, Dani and Cohen-Or, Daniel and Huang, Hui},
  booktitle={Proc. of the Computer Vision and Pattern Recognition Conf.},
  pages={24690--24699},
  year={2025}
}

@inproceedings{lin2025make,
  title={Make me happier: Evoking emotions through image diffusion models},
  author={Lin, Qing and Zhang, Jingfeng and Ong, Yew-Soon and Zhang, Mengmi},
  booktitle={Proc. of the IEEE/CVF International Conf. on Computer Vision},
  pages={16367--16376},
  year={2025}
}

@article{mao2025emoagent,
  title={EmoAgent: Multi-Agent Collaboration of Plan, Edit, and Critic, for Affective Image Manipulation},
  author={Mao, Qi and Hu, Haobo and He, Yujie and Gao, Difei and Chen, Haokun and Jin, Libiao},
  journal={arXiv:2503.11290},
  year={2025}
}

@inproceedings{dang2025emoticrafter,
  title={Emoticrafter: Text-to-emotional-image generation based on valence-arousal model},
  author={Dang, Shengqi and He, Yi and Ling, Long and Qian, Ziqing and Zhao, Nanxuan and Cao, Nan},
  booktitle={Proc. of the IEEE/CVF International Conf. on Computer Vision},
  pages={15218--15228},
  year={2025}
}

@article{jia2025emofeedback2,
  title={EmoFeedback2: Reinforcement of Continuous Emotional Image Generation via LVLM-based Reward and Textual Feedback},
  author={Jia, Jingyang and Shu, Kai and Yang, Gang and Xing, Long and Chen, Xun and Liu, Aiping},
  journal={arXiv:2511.19982},
  year={2025}
}

@article{xia2025muse,
  title={MUSE: Manipulating Unified Framework for Synthesizing Emotions in Images via Test-Time Optimization},
  author={Xia, Yingjie and Wang, Xi and Shi, Jinglei and Kalogeiton, Vicky and Yang, Jian},
  journal={arXiv:2511.21051},
  year={2025}
}

@MISC{JDev2019,
    title =        {News Feed Eradicator},
    author =       {JDev},
    year =         {2019},
    note = {\href{https://chrome.google.com/webstore/detail/news-feed-eradicator/fjcldmjmjhkklehbacihaiopjklihlgg?hl=en}{chrome.google.com/nfe}},
}

@MISC{LessPhone2019,
    title =        {LessPhone},
    author =       {Mohan, Aswin},
    year =         {2019},
    note = {\href{https://play.google.com/store/apps/details?id=me.aswinmohan.nophone&pli=1}{play.google.com/apps/lessphone}},
}

@MISC{Forest2018,
    title =        {Forest - Stay focused, be present},
    author =       {forestapp.cc},
    year =         {2018},
    note = {\href{https://www.forestapp.cc/}{forestapp.cc}},
}

@MISC{Apple2023,
    title =        {Keep track of your screen time on iPhone},
    author =       {Apple},
    year =         {2023},
    note = {\href{https://support.apple.com/guide/iphone/view-your-screen-time-summary-iph24dcd4fb8/ios}{apple.com}},
}

@inproceedings{LyngsHackMyself2020,
author = {Lyngs, Ulrik and Lukoff, Kai and Slovak, Petr and Seymour, William and Webb, Helena and Jirotka, Marina and Zhao, Jun and Van Kleek, Max and Shadbolt, Nigel},
title = { 'I Just Want to Hack Myself to Not Get Distracted': Evaluating Design Interventions for Self-Control on Facebook},
year = {2020},
publisher = {ACM},
address = {New York, NY, USA},
doi = {10.1145/3313831.3376672},
booktitle = {Proc. of the 2020 CHI Conf. on Human Factors in Computing Systems},
numpages = {15},
location = {Honolulu, HI, USA},
series = {CHI '20}
}

@book{brehm2013psychological,
  title={Psychological reactance: A theory of freedom and control},
  author={Brehm, Sharon and Brehm, Jack},
  year={2013},
  publisher={Academic Press}
}

@inproceedings{lukoff2022designing,
  title={Designing to support autonomy and reduce psychological reactance in digital self-Control tools},
  author={Lukoff, Kai and Lyngs, Ulrik and Alberts, Lize},
  booktitle={Position Papers for the Workshop “Self-Determination Theory in HCI: Shaping a Research Agenda” at the Conf. on Human Factors in Computing Systems (CHI’22)},
  volume={5},
  year={2022}
}

@article{lyngs2022goldilocks,
  title={The Goldilocks level of support: Using user reviews, ratings, and installation numbers to investigate digital self-control tools},
  author={Lyngs, Ulrik and Lukoff, Kai and Csuka, Laura and Slov{\'a}k, Petr and Van Kleek, Max and Shadbolt, Nigel},
  journal={International journal of human-computer studies},
  volume={166},
  pages={102869},
  year={2022},
  publisher={Elsevier}
}

@inproceedings{ko2015nugu,
  title={NUGU: a group-based intervention app for improving self-regulation of limiting smartphone use},
  author={Ko, Minsam and Yang, Subin and Lee, Joonwon and Heizmann, Christian and Jeong, Jinyoung and Lee, Uichin and Shin, Daehee and Yatani, Koji and Song, Junehwa and Chung, Kyong-Mee},
  booktitle={Proc. of the 18th ACM conf. on computer supported cooperative work \& social computing},
  pages={1235--1245},
  year={2015}
}

@inproceedings{lu2024interactout,
  title={InteractOut: Leveraging Interaction Proxies as Input Manipulation Strategies for Reducing Smartphone Overuse},
  author={Lu, Tao and Zheng, Hongxiao and Zhang, Tianying and Xu, Xuhai “Orson” and Guo, Anhong},
  booktitle={Proc. of the CHI Conf. on Human Factors in Computing Systems},
  pages={1--19},
  year={2024}
}

@article{hu2018exposure,
  title={Exposure: A white-box photo post-processing framework},
  author={Hu, Yuanming and He, Hao and Xu, Chenxi and Wang, Baoyuan and Lin, Stephen},
  journal={ACM Transactions on Graphics (TOG)},
  volume={37},
  number={2},
  pages={1--17},
  year={2018},
  publisher={ACM New York, NY, USA}
}

@article{dan2011geneva,
  title={The Geneva affective picture database (GAPED): a new 730-picture database focusing on valence and normative significance},
  author={Dan-Glauser, Elise S and Scherer, Klaus R},
  journal={Behavior research methods},
  volume={43},
  pages={468--477},
  year={2011},
  publisher={Springer}
}

@article{lang1997international,
  title={International affective picture system (IAPS): Technical manual and affective ratings},
  author={Lang, Peter J and Bradley, Margaret M and Cuthbert, Bruce N and others},
  journal={NIMH Center for the Study of Emotion and Attention},
  volume={1},
  number={39-58},
  pages={3},
  year={1997},
  publisher={Florida, FL}
}

@inproceedings{radford2021learning,
  title={Learning transferable visual models from natural language supervision},
  author={Radford, Alec and Kim, Jong Wook and Hallacy, Chris and Ramesh, Aditya and Goh, Gabriel and Agarwal, Sandhini and Sastry, Girish and Askell, Amanda and Mishkin, Pamela and Clark, Jack and others},
  booktitle={International conf. on machine learning},
  pages={8748--8763},
  year={2021},
  organization={PMLR}
}

@inproceedings{eriba2019kornia,
  author    = {E. Riba and D. Mishkin and D. Ponsa and E. Rublee and G. Bradski},
  title     = {Kornia: an Open Source Differentiable Computer Vision Library for PyTorch},
  booktitle = {Winter Conf. on Applications of Computer Vision},
  year      = {2020},
  url       = {https://arxiv.org/pdf/1910.02190.pdf}
}

@article{bradley1994measuring,
  title={Measuring emotion: the self-assessment manikin and the semantic differential},
  author={Bradley, Margaret M and Lang, Peter J},
  journal={Journal of behavior therapy and experimental psychiatry},
  volume={25},
  number={1},
  pages={49--59},
  year={1994},
  publisher={Elsevier}
}

@inproceedings{lin2014microsoft,
  title={Microsoft coco: Common objects in context},
  author={Lin, Tsung-Yi and Maire, Michael and Belongie, Serge and Hays, James and Perona, Pietro and Ramanan, Deva and Doll{\'a}r, Piotr and Zitnick, C Lawrence},
  booktitle={Computer Vision--ECCV 2014: 13th European Conf., Zurich, Switzerland, September 6-12, 2014, Proc., Part V 13},
  pages={740--755},
  year={2014},
  organization={Springer}
}

@inproceedings{kim2020multimodal,
  title={Multimodal Post Attentive Profiling for Influencer Marketing},
  author={Kim, Seungbae and Jiang, Jyun-Yu and Nakada, Masaki and Han, Jinyoung and Wang, Wei},
  booktitle={Proc. of The Web Conf. 2020},
  pages={2878--2884},
  year={2020}
}

@article{Gebhardt2025Generative,
  author = {Gebhardt, Christoph and Willardt, Robin and Sadat, Seyedmorteza and Ning, Charlotte and Brombach, Andreas and Song, Jie and Hilliges, Otmar and Holz, Christian},
  title = {Regressor-Guided Image Editing Shifts Emotion and Disengagement Timing in Social Media},
  journal={arXiv:2406.08472},
  year={2025}
}

@article{mood2026,
  title   = {MooD: An Efficient VA-Driven Affective Image Editing Framework via Fine-Grained Semantic Control},
  author  = {Yin, Xinyi and Wang, Yiduo and Hu, Tingqi and Si, Meicong and Shi, Yunyun and Chen, Shi and Wang, Hao and Xue, Junxiao and Wu, Xuecheng},
  journal = {arXiv:2605.02521},
  year    = {2026}
}

@inproceedings{weng2023affective,
  title     = {Affective Image Filter: Reflecting Emotions from Text to Images},
  author    = {Weng, Shuchen and Zhang, Peixuan and Chang, Zheng and Wang, Xinlong and Li, Si and Shi, Boxin},
  booktitle = {IEEE/CVF International Conf. on Computer Vision (ICCV)},
  pages     = {10810--10819},
  year      = {2023}
}

@inproceedings{qin2024mobilenetv4,
  title={MobileNetV4: Universal models for the mobile ecosystem},
  author={Qin, Danfeng and Leichner, Chas and Delakis, Manolis and Fornoni, Marco and Luo, Shixin and Yang, Fan and Wang, Weijun and Banbury, Colby and Ye, Chengxi and Akin, Berkin and others},
  booktitle={European conf. on computer vision},
  pages={78--96},
  year={2024},
  organization={Springer}
}

@misc{timm2024mobilenetv4,
  author       = {{timm}},
  title        = {Model card for mobilenetv4\_conv\_small.e2400\_r224\_in1k},
  howpublished = {\url{https://huggingface.co/timm/mobilenetv4_conv_small.e2400_r224_in1k}},
  year         = {2024}
}

@inproceedings{perez2018film,
  title={Film: Visual reasoning with a general conditioning layer},
  author={Perez, Ethan and Strub, Florian and De Vries, Harm and Dumoulin, Vincent and Courville, Aaron},
  booktitle={Proc. of the AAAI conf. on artificial intelligence},
  volume={32},
  number={1},
  year={2018}
}

@article{dumoulin2018feature,
  title={Feature-wise transformations},
  author={Dumoulin, Vincent and Perez, Ethan and Schucher, Nathan and Strub, Florian and Vries, Harm de and Courville, Aaron and Bengio, Yoshua},
  journal={Distill},
  volume={3},
  number={7},
  pages={e11},
  year={2018}
}

@article{kingma2014adam,
  title={Adam: A method for stochastic optimization},
  author={Kingma, Diederik P and Ba, Jimmy},
  journal={arXiv:1412.6980},
  year={2014}
}

@misc{onnxruntimeweb,
  author       = {{ONNX Runtime}},
  title        = {Build for Web},
  howpublished = {\url{https://onnxruntime.ai/docs/build/web.html}},
  year         = {2026}
}

@misc{mdn_requestvideoframecallback,
  title        = {HTMLVideoElement: requestVideoFrameCallback() method - Web APIs | MDN},
  author       = {{MDN Contributors}},
  howpublished = {\url{https://developer.mozilla.org/en-US/docs/Web/API/HTMLVideoElement/requestVideoFrameCallback}},
  organization = {Mozilla}
}

@article{elliot2019historically,
  title={A historically based review of empirical work on color and psychological functioning: Content, methods, and recommendations for future research},
  author={Elliot, Andrew J},
  journal={Review of General Psychology},
  volume={23},
  number={2},
  pages={177--200},
  year={2019},
  publisher={Sage Publications Sage CA: Los Angeles, CA}
}

@article{elliot2015color,
  title={Color and psychological functioning: a review of theoretical and empirical work},
  author={Elliot, Andrew J},
  journal={Frontiers in psychology},
  volume={6},
  pages={},
  year={2015},
  publisher={Frontiers}
}

@article{yin2024coloredit,
  title={ColorEdit: Training-free image-guided color editing with diffusion model},
  author={Yin, Xingxi and Li, Zhi and Zhang, Jingfeng and Li, Chenglin and Zhang, Yin},
  journal={arXiv:2411.10232},
  year={2024}
}

@article{basu2023editval,
  title={Editval: Benchmarking diffusion based text-guided image editing methods},
  author={Basu, Samyadeep and Saberi, Mehrdad and Bhardwaj, Shweta and Chegini, Atoosa Malemir and Massiceti, Daniela and Sanjabi, Maziar and Hu, Shell Xu and Feizi, Soheil},
  journal={arXiv:2310.02426},
  year={2023}
}

@inproceedings{caron2021emerging,
  title={Emerging properties in self-supervised vision transformers},
  author={Caron, Mathilde and Touvron, Hugo and Misra, Ishan and J{\'e}gou, Herv{\'e} and Mairal, Julien and Bojanowski, Piotr and Joulin, Armand},
  booktitle={2021 IEEE/CVF international conf. on computer vision (ICCV)},
  pages={9630--9640},
  year={2021},
  organization={IEEE}
}
}

\appendix

\section{Method Details}

\label{apx:method}
\begin{figure*}[!t]
    \centering
    \includegraphics[width=0.7\linewidth]{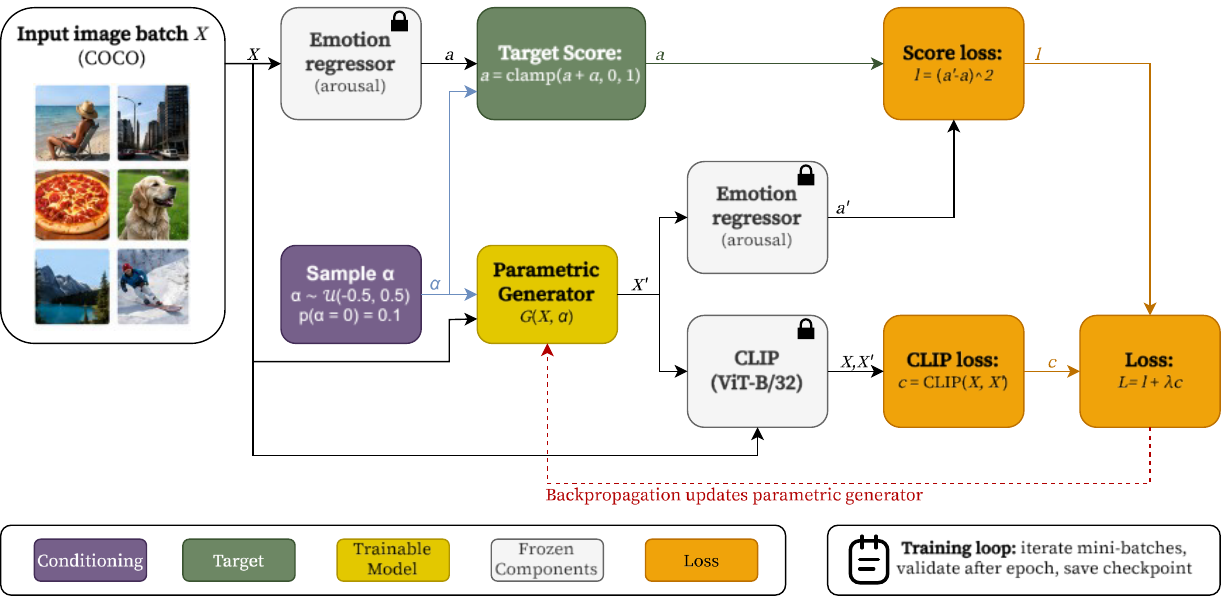}
    \caption{\textbf{Training procedure.} Solid arrows show the forward pass: a sampled conditioning value $\alpha$ and the frozen regressor's estimate of $I$ define the target score $a' = \mathrm{clip}(R_a(I){+}\alpha,0,1)$; the generator $f_\theta$ predicts transform parameters $p$, producing $\hat{I} = T(I,p)$, which is re-scored by $R_a$ and compared against $a'$ ($\mathcal{L}_{\text{affect}}$), while a frozen CLIP encoder supplies a semantic-preservation term ($\mathcal{L}_{\text{CLIP}}$) against the original image. 
    The dashed red arrow shows the backward pass: gradients of the combined objective $\mathcal{L}(\theta)$ propagate back through $R_a$, the CLIP encoder, and $T$ into $f_\theta$; $R_a$ and the CLIP encoder (boxes with locks) remain frozen throughout.}
    \label{fig:training-process-appendix}
\end{figure*}

\subsection{Full Parameterization of the Transform $T$}
\label{apx:transform}

We follow Gebhardt et al.'s \cite{Gebhardt2025Generative} formulation of $T$, implemented with standard differentiable Kornia~\cite{eriba2019kornia} operations for exposure, saturation, contrast, sharpening, and blur. The tone curve follows Mejjati et al.~\cite{mejjati2020look} and the color curve follows Hu et al.~\cite{hu2018exposure}; both are applied as $K$-step piecewise functions. \cref{tab:transform-params} lists the full parameterization of $T$.

\begin{table}[b]
\centering
\small
\begin{tabular}{@{}lllll@{}}
\toprule
Operation & Dim. & Raw Range & Range $[m,M]$ & Identity \\
\midrule
Exposure    & 1    & $(-\infty,\infty)$ & $[-0.1, 1]$   & 0.0 \\
Saturation  & 1    & $[0,\infty)$        & $[0.5, 3]$    & 1.0 \\
Tone curve  & $K$  & $(-\infty,\infty)$  & $[0, 3]$      & 1.0 \\
Color curve & $3K$ & $(-\infty,\infty)$  & $[0, 3]$      & 1.0 \\
Contrast    & 1    & $[0,\infty)$        & $[0.5, 3]$    & 1.0 \\
Sharpness   & 1    & $[0,\infty)$        & $[0, 3]$      & 1.0 \\
Blur ($\sigma$) & 1 & $(0,\infty)$      & $(0, 3)$      & $\approx 0$ \\
\bottomrule
\end{tabular}
\caption{\textbf{Sub-transforms composing $T$.} Raw range is the domain of the sub-transform; range is the interval $[m,M]$ enforced by $f_\mathrm{sig}$ (\cref{eq:param-map}) on the values predicted by $f_\theta$. Parameters are applied in the fixed order listed above.}
\label{tab:transform-params}
\end{table}

$T$ applies these seven operations sequentially in the order listed in \cref{tab:transform-params}: exposure (exponential brightness scaling), saturation (scales color intensity in HSV space), an 8-step piecewise tone curve, an 8-step per-channel color curve, contrast (scales deviation from mean intensity), sharpening (high-frequency edge amplification), and Gaussian blur. Tone and color curves are multi-dimensional sub-transforms ($K{=}8$); all others are scalar.

\paragraph{Parameter mapping.} Raw head outputs are mapped before being consumed by $T$. We evaluate three variants: no mapping, a sigmoid-based mapping,
\begin{equation}
f_{\text{sig}}(u, m, M, i) = m + (M - m)\,\sigma\!\left(u + \log\frac{i - m}{M - i}\right),
\label{eq:param-map}
\end{equation}
and a $\tanh$-based mapping that treats the regions above and below $i$ asymmetrically,
\begin{equation}
f_{\tanh}(u, m, M, i, s) =
\begin{cases}
i + (M - i)\tanh(u / s), & u \ge 0, \\
i + (i - m)\tanh(u / s), & u < 0.
\end{cases}
\label{eq:param-map-tanh}
\end{equation}
Here $u \in \mathbb{R}$ is the unconstrained raw output of a prediction head, $[m, M]$ is the target parameter's allowed range (\cref{tab:transform-params}), $i$ is that parameter's identity value, and $s$ is a scale constant controlling how sharply $f_{\tanh}$ saturates toward its bounds.
Both mappings guarantee $f(0) = i$, so an untrained or zero-conditioned generator reduces to the identity transform. 
\cref{apx:ablation-curves} presents an ablation study over the choice of mapping function.

\begin{table}[b]
    \centering
    \small
    \begin{tabular}{ll}
        \toprule
        \textbf{Setting} & \textbf{Value} \\
        \midrule
        Optimizer          & Adam~\cite{kingma2014adam}\\
        Learning rate       & $1\times10^{-4}$ \\
        Batch size          & 32 (4 $\times$ 8 GPUs) \\
        Hardware            & 8 $\times$ NVIDIA Tesla P100 \\
        Dataset              & MS-COCO~\cite{lin2014microsoft} \\
        $\lambda_{\mathrm{clip}}$ & 0.05 \\
        $\alpha$ sampling     & $\mathcal{U}(-0.5, 0.5)$, clipped \\
        \bottomrule
    \end{tabular}
    \caption{\textbf{Training configuration for the parametric generator.}}
    \label{tab:training}
\end{table}

\subsection{Training Procedure}
\label{apx:training-figure}

\cref{fig:training-process-appendix} details the full training loop introduced in \cref{sec:method:training}, including the conditioning pipeline and gradient flow. For each input image $I$, a conditioning value $\alpha$ is sampled uniformly from $[-0.5, 0.5]$. A frozen, pretrained affect regressor $R$ estimates the original arousal score $R_a(I)$, and the target score is obtained as $a' = \mathrm{clip}(R_a(I) + \alpha,\, 0, 1)$. 
The conditional parametric generator $f_\theta$ takes $I$ and $\alpha$ as input and predicts the parameters $p$ of the global transformation $T$, producing the edited image $\hat{I} = T(I, p)$. 
This edited image is passed through the same frozen regressor $R$ a second time, and $f_\theta$ is optimized so that $R_a(\hat{I})$ matches the target score $a'$ (\cref{eq:loss}). 
A frozen CLIP image encoder additionally supplies a semantic-preservation term that penalizes drift between $I$ and $\hat{I}$ in embedding space (\cref{eq:clip}). 
Gradients from the combined objective flow back through $T$ and into $f_\theta$ only: $R$ and the CLIP encoder are frozen throughout and receive no parameter updates. \cref{tab:training} summarizes the training configuration.

\subsection{Affect Regressor}
\label{apx:va-dataset}
The frozen affect regressor $R$ is trained by Gebhardt et al.~\cite{Gebhardt2025Generative} on a dataset (20{,}460 images) integrating ten open-science affective image databases, providing human-rated valence and arousal annotations consistent with the CMA. Their integrated dataset exhibits a V-shaped relationship between valence and arousal: images rated toward either extreme of valence tend to also be rated high in arousal, while neutral-valence images cluster toward lower arousal (\cite{Gebhardt2025Generative}; Fig. 10). This pattern is consistent with the population-level valence--arousal coupling visible in emotional image databases~\cite{lang1997international,dan2011geneva}.
This coupling is referenced to contextualize the indirect effect of our arousal-only conditioning on perceived valence (\cref{sec:user-study:results}): since $R$ is trained on data where valence and arousal are structurally entangled at the population level, an arousal-directed edit is compatible with a valence side-effect.

\section{System Details}
\label{apx:system}

\subsection{Implementation Constraints}
\label{sec:system:decisions}
We report implementation choices that may generalize to other in-browser, on-device inference systems.

\paragraph{Applying the transform.}
Executing $T$ inside the inference runtime is prohibitively slow: an ONNX-based implementation measures $55.1$\,ms mean latency per frame, against $0.5$\,ms for the GLSL fragment shader under WebGL2 (AMD Ryzen 7 5700X3D / RTX 4070 Ti Super). Real-time rendering is therefore feasible only if parameter prediction and parameter application are separated across runtimes, with $T$ evaluated on the GPU via shader.

\paragraph{Inference runtime.}
Among web runtimes, only the CPU/WASM backend of ONNX Runtime Web~\cite{onnxruntimeweb} provides full operator coverage; GPU-backed backends (WebGL, WebGPU, WebNN) currently implement subsets. Backend choice thus constrains the admissible operator set of the model itself, and we treat full coverage as a hard requirement. ONNX additionally exports directly from our PyTorch pipeline, avoiding a separate conversion path.

\paragraph{Frame access.}
Short-form video can be intercepted via screen capture, platform media APIs, or at the DOM level through a WebView. We use the WebView path, which requires no elevated permissions and operates on standard DOM video elements, making it portable across web-based front-ends. It is correspondingly coupled to the DOM of the target application and breaks if its structure changes.

\section{Evaluation Details}

\subsection{Model}
\label{apx:model-evaluation}

\subsubsection{Architecture Ablation}
\label{apx:ablation-curves}

\begin{figure}[t]
    \centering
    \centering
    \begin{subfigure}[b]{\linewidth}
        \centering
        \includegraphics[width=\linewidth]{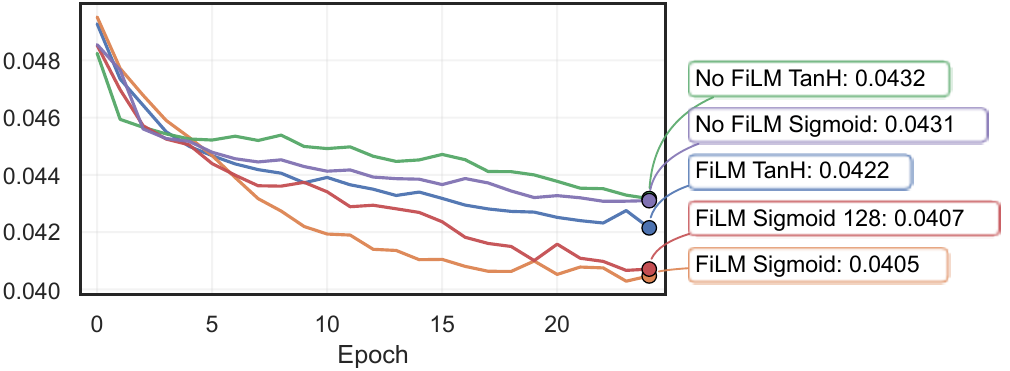}
        \caption{Overall loss.}
        \label{fig:loss-architecture}
    \end{subfigure}
    \begin{subfigure}[b]{\linewidth}
        \centering
        \includegraphics[width=\linewidth]{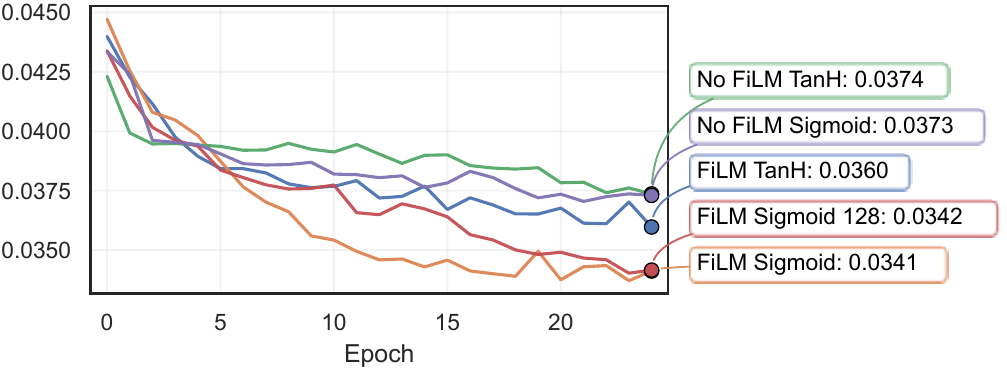}
        \caption{Arousal MSE term.}
        \label{fig:arousal-architecture}
    \end{subfigure}
    \begin{subfigure}[b]{\linewidth}
        \centering
        \includegraphics[width=\linewidth]{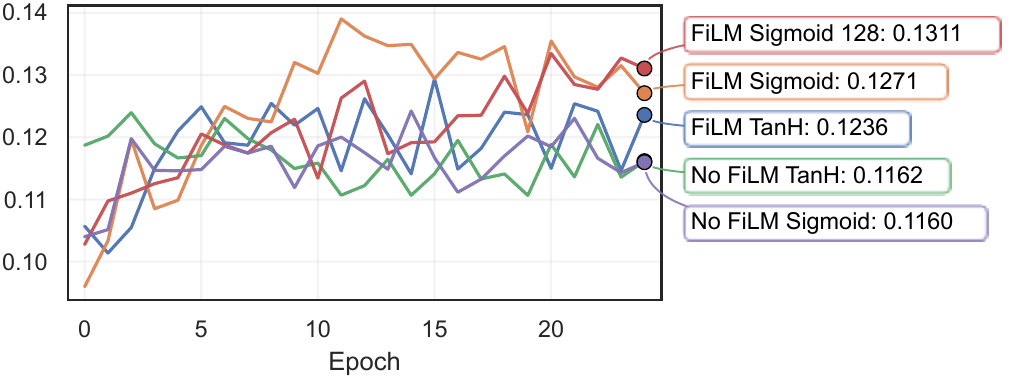}
        \caption{CLIP loss.}
        \label{fig:clip-architecture}
    \end{subfigure}
    \caption{\textbf{Full training curves for the FiLM ablation.}
    Validation overall loss, arousal MSE term, and CLIP preservation loss over training, for all five configurations described in \cref{apx:ablation-curves}.}
    \label{fig:training-curves-appendix}
\end{figure}

\cref{fig:training-curves-appendix} shows the validation curves underlying the FiLM ablation summarized in \cref{sec:evaluation:model}. We compare five configurations: a dense trunk with FiLM residual blocks and a sigmoid parameter mapping (\textit{FiLM Sigmoid}), the same architecture with trunk width increased from 64 to 128 (\textit{FiLM Sigmoid 128}), FiLM blocks with a $\tanh$ mapping (\textit{FiLM TanH}), and two ablations that remove the FiLM blocks entirely, differing only in the head mapping (\textit{No FiLM Sigmoid}, \textit{No FiLM TanH}). All five share the same trunk depth (3) and head depth/width (2, 64). Each is tracked on three validation metrics: overall training loss (\cref{eq:loss}), the arousal MSE term of \cref{eq:loss}, and CLIP loss (\cref{eq:clip}).

The two sigmoid-mapped FiLM configurations achieve near-identical, and the lowest, overall loss and arousal MSE term among all five variants; widening the trunk from 64 to 128 yields no measurable improvement over the base width (64), indicating the additional capacity is not needed at this task scale. \textit{FiLM TanH} and both no-FiLM variants trail on these three metrics. The pattern reverses for CLIP loss: the no-FiLM variants obtain the lowest (best) CLIP loss, \textit{FiLM TanH} is intermediate, and both sigmoid FiLM variants obtain the highest CLIP loss. This is the accuracy preservation trade-off referenced in the main text: FiLM's stronger conditioning improves affect-score fidelity at some cost to semantic preservation. We select \textit{FiLM Sigmoid} (64 width) for the final model, as it matches the wider variant on affect fidelity at lower computational costs.

\subsubsection{Loss Ablation}
\label{apx:clip-ablation-curves}
\begin{figure}[t]
    \centering
    \begin{subfigure}[b]{\linewidth}
        \centering
        \includegraphics[width=0.82\linewidth]{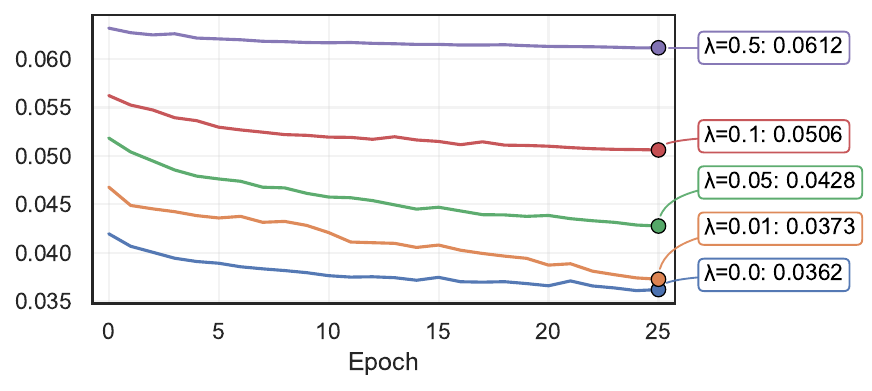}
        \caption{Overall loss.}
        \label{fig:clip-ablation-loss}
    \end{subfigure}
    \begin{subfigure}[b]{\linewidth}
        \centering
        \includegraphics[width=0.82\linewidth]{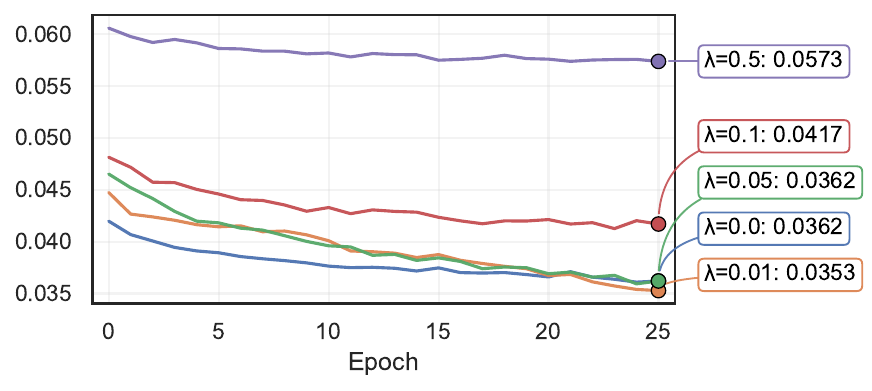}
        \caption{Arousal MSE term.}
        \label{fig:clip-ablation-arousal}
    \end{subfigure}
    \begin{subfigure}[b]{\linewidth}
        \centering
        \includegraphics[width=0.82\linewidth]{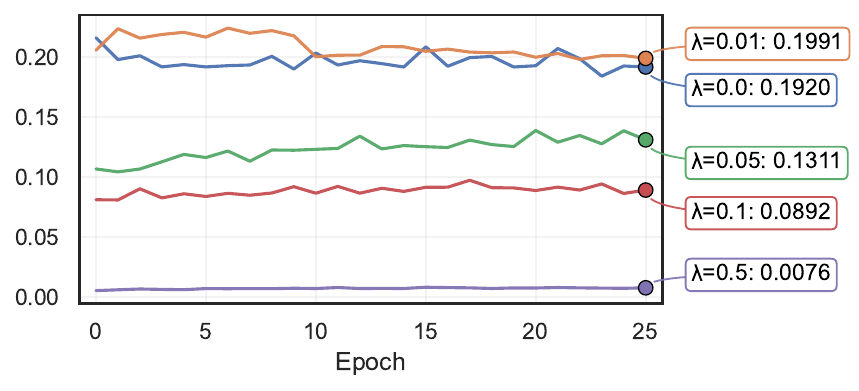}
        \caption{CLIP loss.}
        \label{fig:clip-ablation-clip}
    \end{subfigure}
    \caption{\textbf{Full training curves for the $\lambda_{\mathrm{clip}}$ ablation.} Validation overall loss, arousal MSE term, and CLIP preservation loss over training, for all five $\lambda_{\mathrm{clip}}$ values described in \cref{apx:clip-ablation-curves}.}
    \label{fig:clip-ablation-curves-appendix}
\end{figure}

\cref{fig:clip-ablation-curves-appendix} shows the full validation curves underlying the $\lambda_{\mathrm{clip}}$ ablation summarized in \cref{sec:evaluation:model}, tracked across the same three metrics as the architecture ablation.

The trade-off is stable throughout training, not only at the final checkpoint: arousal MSE and CLIP loss separate by $\lambda_{\mathrm{clip}}$ from early epochs onward and maintain a consistent order across all 25 epochs. 
Arousal MSE curves for $\lambda_{\mathrm{clip}} \in \{0, 0.01, 0.05\}$ converge closely by the final epochs, while $\lambda_{\mathrm{clip}}=0.1$ and $0.5$ remain clearly separated throughout.

\begin{table*}[t]
\setlength{\tabcolsep}{5pt}
\centering
\small
\begin{tabular}{@{}p{0.13\linewidth}p{0.42\linewidth}cc l@{}}
\toprule
& & \multicolumn{3}{c}{Prior literature} \\
\cmidrule(l){3-5}
Attribute & Computation & Valence & Arousal & References \\
\midrule
Brightness & Mean grayscale pixel intensity. & $(+)$ & -- & \cite{specker2018universal, gao2007analysis} \\
Saturation & Mean of the HSV saturation channel. & $(+)$ & $(+)$ & \cite{simola2015affective, redies2020global} \\
Contrast & Std.\ dev.\ of grayscale pixel intensities. & $(+)$ & -- & \cite{yang2020emotion} \\
Colorfulness & Std.\ dev.\ plus weighted mean of distances from the mean in the LAB $a$/$b$ channels. & $(+)/(-)$\textsuperscript{a} & $(+)$ & \cite{bekhtereva2017bringing, cano2009affective} \\
Blur & Laplacian variance of grayscale image (high $=$ less blur). & $(-)$ & $(-)$ & \cite{de2010effects} \\
Lighting div. & Std.\ dev.\ of the LAB lightness ($L$) channel. & -- & $(+)$ & \cite{delplanque2007spatial} \\
\bottomrule
\multicolumn{5}{@{}p{0.95\linewidth}@{}}{\footnotesize\textsuperscript{a}Colorfulness amplifies valence in both directions rather than shifting it monotonically.}
\end{tabular}
\caption{\textbf{Image attribute metrics and their affective associations.}
Associations are reported from prior literature, where $(+)$/$(-)$ denote a positive/negative association and ``--'' indicates none reported.}
\label{tab:attribute-metrics}
\end{table*}

\subsubsection{Arousal Steering} 
\label{apx:linear-control}
To further quantify control fidelity, we fit an ordinary least-squares model regressing the regressor's output on $\alpha$, evaluated at $17$ evenly spaced values $\alpha \in [-0.5, 0.5]$ with step size $0.0625$:
\begin{equation}
R_a\bigl(T(I, f_\theta(I, \alpha))\bigr) = \beta_1 \alpha + \beta_0 + \varepsilon,
\label{eq:ols-control-appendix}
\end{equation}

Model evaluation was conducted on a fixed set of 5{,}000 images from the COCO~2017 test set~\cite{lin2014microsoft}, rather than the paired 500-image subset used for the baseline comparison in \cref{sec:evaluation:model}; slopes are therefore not directly comparable across the two analyses.

\begin{figure}[b]
    \centering
    \begin{subfigure}[b]{0.495\linewidth}
        \centering
        \includegraphics[width=\linewidth]{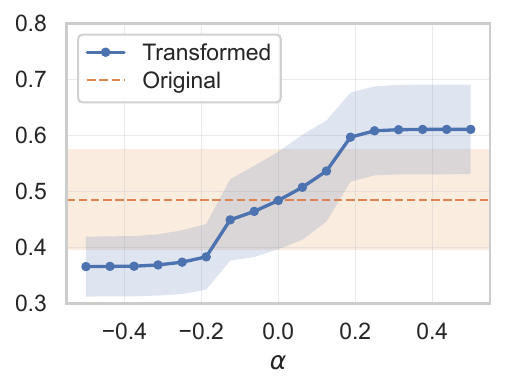}
        \caption{Transformed arousal.}
        \label{fig:ols-arousal}
    \end{subfigure}
    \begin{subfigure}[b]{0.495\linewidth}
        \centering
        \includegraphics[width=\linewidth]{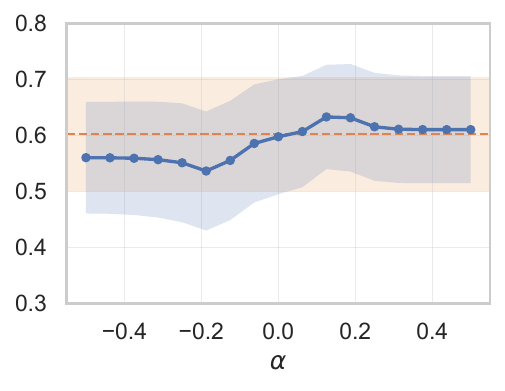}
        \caption{Transformed valence.}
        \label{fig:ols-valence}
    \end{subfigure}
    \caption{\textbf{Regressor-predicted arousal and valence as a function of $\alpha$.} (a) Arousal increases monotonically with $\alpha$ and saturates beyond $|\alpha|\gtrsim0.2$. (b) Valence, not directly controlled, shows similar turning points, indicating an arousal influence.}
    \label{fig:ols}
\end{figure}

The fitted slope is $\beta_1 = 0.317$ ($p < .001$), confirming a statistically significant, approximately linear relationship between the conditioning value and the induced change in predicted arousal; a slope below 1 indicates the edit systematically underestimates the requested effect size.
The intercept $\beta_0 = 0.006$ ($p < .001$) is small but significant, indicating a slight bias even at $\alpha = 0$.
The fit is moderate ($R^2 = 0.668$; MSE $= 0.0341$), with the remaining variance plausibly explained by the non-linear, image-dependent response detailed in \cref{apx:preservation}.
Consistent with that non-linearity, the arousal--$\alpha$ relationship is sigmoidal rather than linear (\cref{fig:ols-arousal}), saturating for $|\alpha| \gtrsim 0.2$, plausibly because low-level transformations can shift perceived arousal only so far without altering image content.

We repeat the same OLS fit with predicted valence as the response variable ($R_v$; \cref{fig:ols-valence}). 
The linear fit is weak ($R^2 = 0.091$); the slope is small but significant and positive ($\beta_1 = 0.089$, $p < .001$). The response is non-monotonic rather than flat: valence dips to a local minimum near $\alpha \approx -0.2$ and rises to a local maximum near $\alpha \approx 0.2$, coinciding with the arousal curve's transition region. 
The effect of arousal control on valence is consistent with the valence--arousal coupling observed in affective datasets (\cref{apx:va-dataset}).

\begin{figure}[b]
    \centering
    \includegraphics[width=\linewidth]{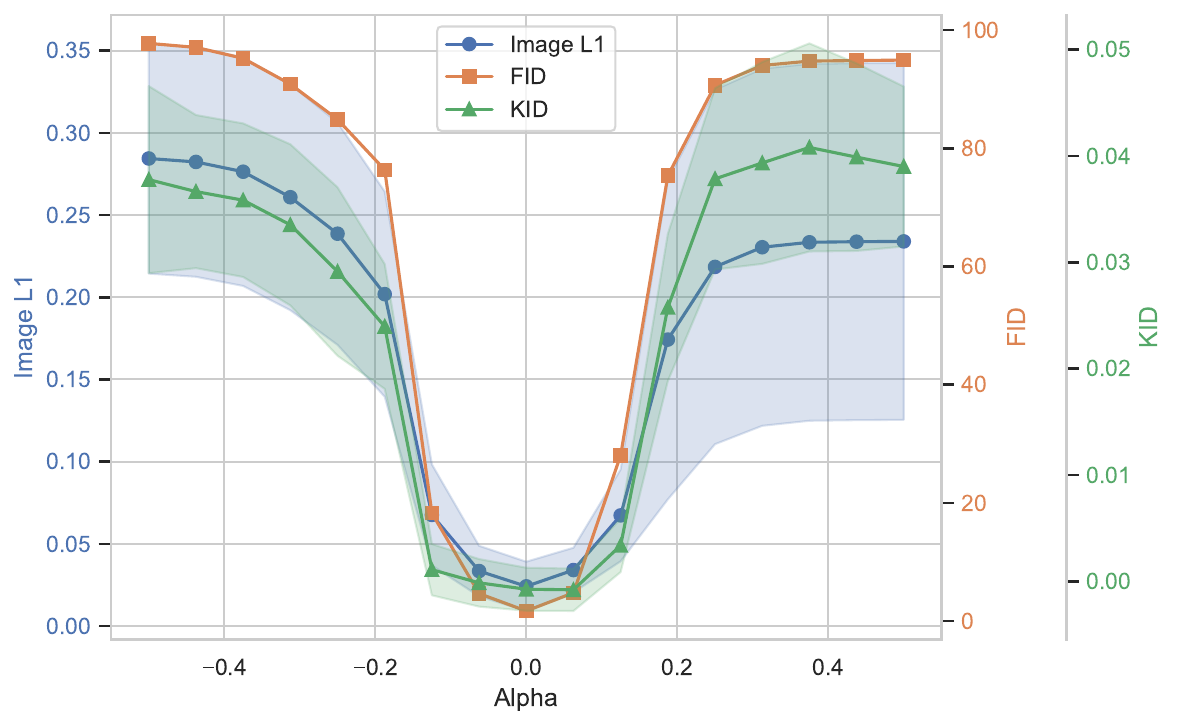}
    \caption{\textbf{Preservation metrics over $\alpha$.} FID, KID and L1 reconstruction error as a function of the conditioning value $\alpha$, showing the parabolic increase near $\alpha=0$ followed by a sharp transition and subsequent flattening.}
    \label{fig:preservation-appendix}
\end{figure}

\subsubsection{Preservation Metrics}
\label{apx:preservation}

\cref{fig:preservation-appendix} extends the DINO results summarized in \cref{sec:evaluation:model} by plotting FID, KID, and L1 individually as a function of the conditioning value $\alpha$, evaluated at 17 fixed values in $[-0.5, 0.5]$ over 1{,}000 images from the COCO 2017 test set. All three metrics are approximately symmetric around $\alpha=0$ and increase with $|\alpha|$, consistent with $\alpha$ controlling edit strength. The increase is not linear: near the identity setting, error grows in an approximately quadratic (parabolic) fashion, after which it exhibits a sharp jump to a regime of much stronger modification, and finally flattens at the largest $|\alpha|$ magnitudes, indicating that further increases in $\alpha$ no longer translate into proportionally stronger edits. The variance of L1 and KID across images grows alongside their means, showing that the strength of the edit induced by a given $\alpha$ is not uniform across images. Some images are edited more strongly than others at the same nominal conditioning strength.
We omit FID variances, as \emph{torch-fidelity}, the package used for computation, does not provide them.

\begin{figure}[t]
    \centering
    \includegraphics[width=\linewidth]{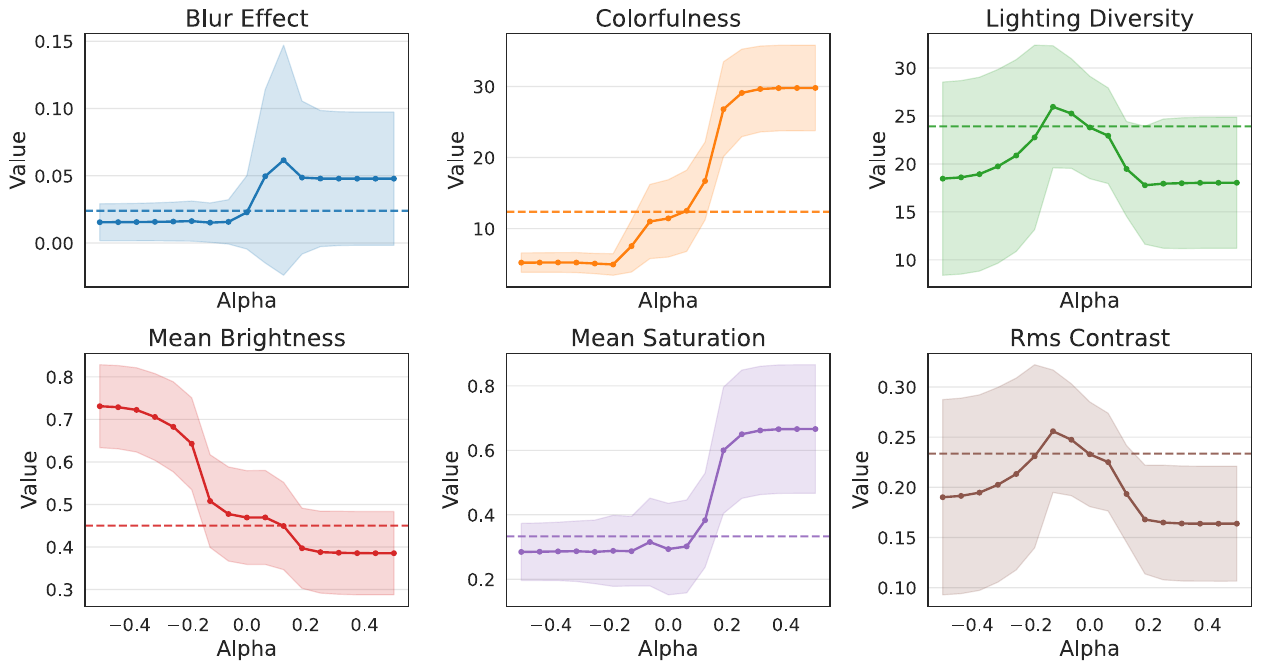}
    \caption{\textbf{Evolution of image attributes over $\alpha$.} All six attributes from \cref{tab:attribute-metrics}, plotted against the conditioning value $\alpha$.}
    \label{fig:attributes-appendix}
\end{figure}

\begin{table}[b]
    \centering
    \small
    \begin{tabular}{lrrrrr}
        \toprule
        \textbf{Method} & \textbf{Median} & \textbf{Mean} & \textbf{p95} & \textbf{Min} & \textbf{Max} \\
        \midrule
        Ours (ms) & 3.71 & 3.72  & 4.34  & 3.05 & 4.99 \\
        \cite{Gebhardt2025Generative} (s) & 80.92 & 80.71 & 83.01 & 78.81 & 83.10 \\
        \bottomrule
    \end{tabular}
    \caption{\textbf{Per-frame latency on training hardware.} Ours is a forward pass ($n=5{,}000$); \cite{Gebhardt2025Generative} is a 300-step edit ($n=250$).}
    \label{tab:model-latency}
\end{table}

\subsubsection{Image Attribute Metrics}
\label{apx:attributes}

\paragraph{Attribute metrics.} \cref{tab:attribute-metrics} lists the image attributes used to characterize edits, and summarizes their established association with valence and/or arousal in the affective-imagery literature, which motivates using them as a diagnostic of whether $\alpha$-conditioning acts along its intended axis. 
The metrics were computed over the same 1{,}000 images of the COCO 2017 test set as the preservation metrics.

\paragraph{Evolution over $\alpha$.} \cref{fig:attributes-appendix} plots all six attributes as a function of $\alpha$.
Colorfulness, linked to arousal in \cref{tab:attribute-metrics}, shifts in the expected direction: it increases with positive $\alpha$ and decreases with negative $\alpha$, supporting $\alpha$ acting along its intended axis.
Saturation, also linked to arousal, instead follows a step function in the theoretically assumed direction: it remains near baseline throughout the negative range and rises sharply only once $\alpha \gtrsim 0.1$, consistent with a threshold rather than a linear response.
Blur follows the same step-function pattern and moves in the expected direction: higher requested arousal should yield sharper, less blurred output, and the metric confirms this, but only for positive $\alpha$; negative $\alpha$ produces no corresponding increase in blur.
Lighting diversity is non-monotonic, peaking near $\alpha = 0$ and decreasing at both extremes, which only partially matches the literature's link between low-frequency luminance variation and arousal.
Valence-linked attributes behave differently but not uniformly: brightness decreases monotonically across the full range, which is explained by the valence–arousal coupling (\cref{apx:va-dataset}) only on the positive-valence side.
Contrast, with no arousal- but a bidirectional valence association, peaks near $\alpha \approx -0.15$ and declines on both sides, a pattern no association predicts.

\begin{table}[t]
    \centering
    \setlength{\tabcolsep}{4pt}
    \small
    \begin{tabular}{lrrrrr}
        \toprule
        \textbf{Stage} & \textbf{Median} & \textbf{Mean} & \textbf{p95} & \textbf{Min} & \textbf{Max} \\
        \midrule
        Copy to canvas & 0.00  & 0.05  & 0.10  & 0.00  & 3.30 \\
        Extract pixels & 7.90  & 8.98  & 17.00 & 2.70  & 42.20 \\
        Inference      & 37.90 & 40.48 & 61.60 & 21.80 & 114.50 \\
        \textbf{Total} & \textbf{46.00} & 49.52 & 77.80 & 24.60 & 138.50 \\
        \bottomrule
    \end{tabular}
    \caption{\textbf{Per-frame latency (ms) on the target device.} Computed on a Samsung Galaxy S23 using $n=1{,}500$ frames.}
    \label{tab:ondevice}
\end{table}

\subsection{System}

\subsubsection{Latency Distributions}
\label{apx:latency}
 
\cref{tab:model-latency} and \cref{tab:ondevice} report the full latency distributions underlying the figures quoted in \cref{sec:evaluation:system}.

\cref{tab:model-latency} compares our approach against the parametric optimization of~\cite{Gebhardt2025Generative} on the same hardware (1$\times$ NVIDIA Tesla P100-SXM2-16GB). We use the authors' implementation and protocol, differing in the arousal term of the objective, the image set, and resolution (both $480{\times}480$).
Our method is measured at batch size 8; the baseline has no batch dimension, being an iterative per-image optimizer.

\cref{tab:ondevice} reports per-frame latency (ms) on the target device (Samsung Galaxy S23), broken down into three measured stages (canvas copy, pixel extraction, and preprocessing/inference) plus their combined total.

\begin{table}[b]
\centering
\small
\begin{tabular}{lrrrr}
\toprule
Component & Mean $\Delta$ (mAh) & SD \\
\midrule
Total     & +31.86 & 10.65 \\
CPU       & +25.13 & 13.78 \\
Screen    & +5.74  &  8.73 \\
Video     & +0.24  &  0.65 \\
WiFi      & $-$0.00 & 0.02 \\
\bottomrule
\end{tabular}
\caption{\textbf{On-device energy cost.} Paired delta (filter-on minus filter-off) in app-attributed power use, per hardware component, across five matched 30-minute sessions on a Samsung Galaxy S23.}
\label{tab:energy}
\end{table}

\subsubsection{On-Device Energy Cost}
\label{apx:energy}

\cref{tab:energy} reports the on-device energy cost of the filter, measured via \texttt{adb shell dumpsys batterystats} across five matched filter-on/filter-off session pairs on the target device (Samsung Galaxy S23). The CPU component shows the clearest attributable cost, consistent with the overhead of on-device model inference (\cref{sec:system:pipeline}). 
The screen component, while positive on average, varies considerably in sign across pairs (range: $-0.3$ to $+19.8$\,mAh). 
Video decode and network draw are unaffected, as expected, since neither is altered by our transform.
Given the small sample ($n{=}5$), we treat these figures as indicative rather than conclusive.

\balance
\section{User Study Details}
\label{apx:user-study}
This study evaluates whether our approaches reduce perceived emotional arousal while maintaining high perceived image quality. 
We compare the adapted images against the original images and their grayscale variant.

\subsection{Study design}
\label{apx:user-study-design}
In the following, we detail conditions, stimuli, task and participants of the experiment.
Our study design (task, metrics, and hypotheses) follows a prior study~\cite{Gebhardt2025Generative}, differing in the stimuli and in three of the five conditions.
We also drop image observation time as a dependent variable, as single-exposure viewing duration is not expected to be sensitive to the manipulation in this setting.

\paragraph{Conditions}
The study contains one within-subject factor, \ivfilter{}, with 5 factor levels: \cdefault{}, \cgray{}, \cone{} ($\alpha = -0.11$), \ctwo{} ($\alpha = -0.12$), \cthree{} ($\alpha = -0.13$).
We selected these $\alpha$-values based on the identified operating range in the technical evaluation, balancing the trade-off between reducing evoked arousal and maintaining fidelity to the original images.

Each of the five conditions was applied to the same block of 12 stimulus images (see \emph{Stimuli}), presented to participants in a fixed order.
To avoid order effects, we used Latin-square counterbalancing to adjust the order of conditions between participants.

\begin{figure*}[t]
  \centering
  \begin{subfigure}[t]{0.32\linewidth}
    \centering
    \includegraphics[width=\linewidth]{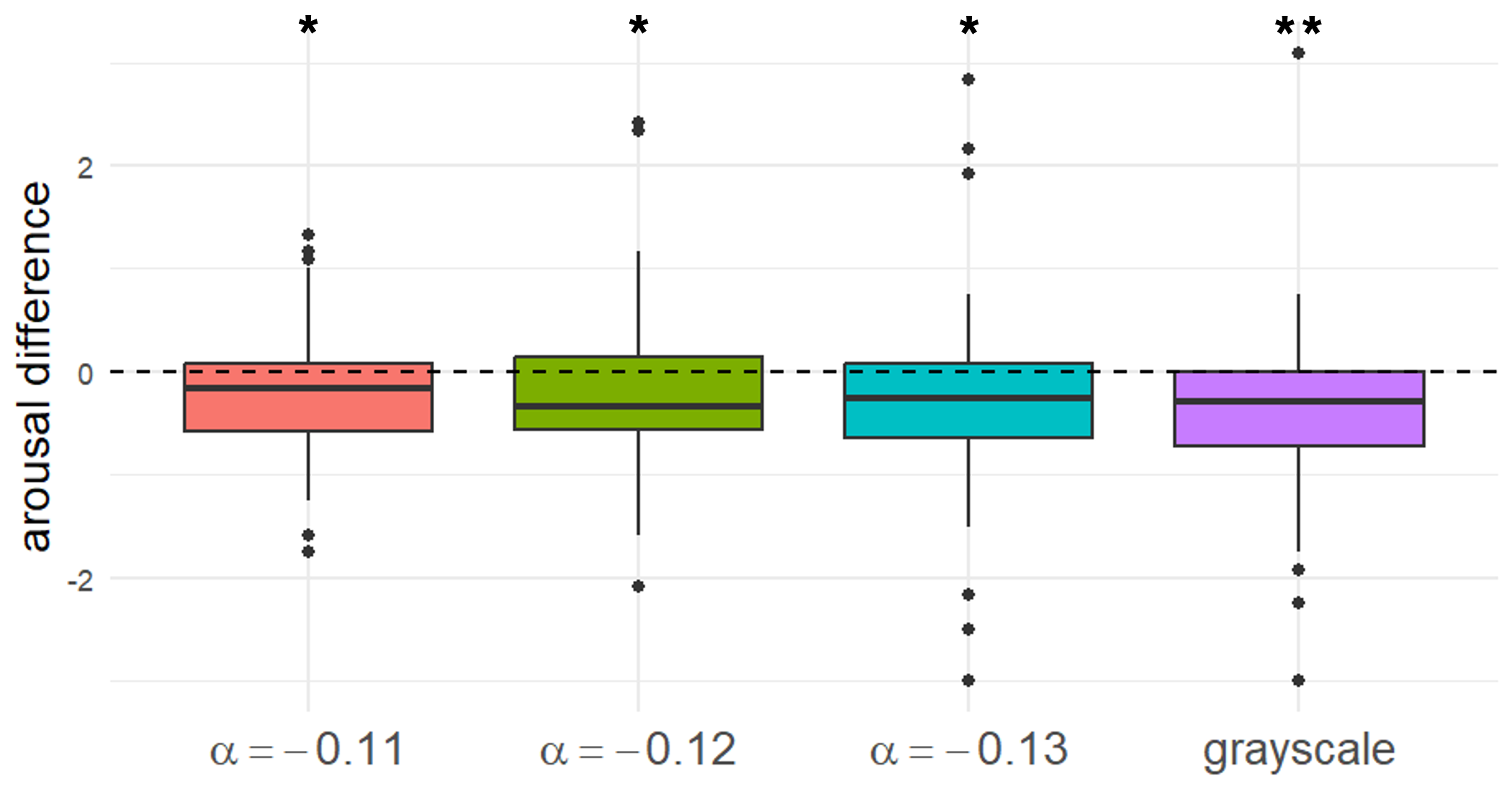}
    \caption{}
    \label{fig:diff-ar}
  \end{subfigure}
  \hfill
  \begin{subfigure}[t]{0.32\linewidth}
    \centering
    \includegraphics[width=\linewidth]{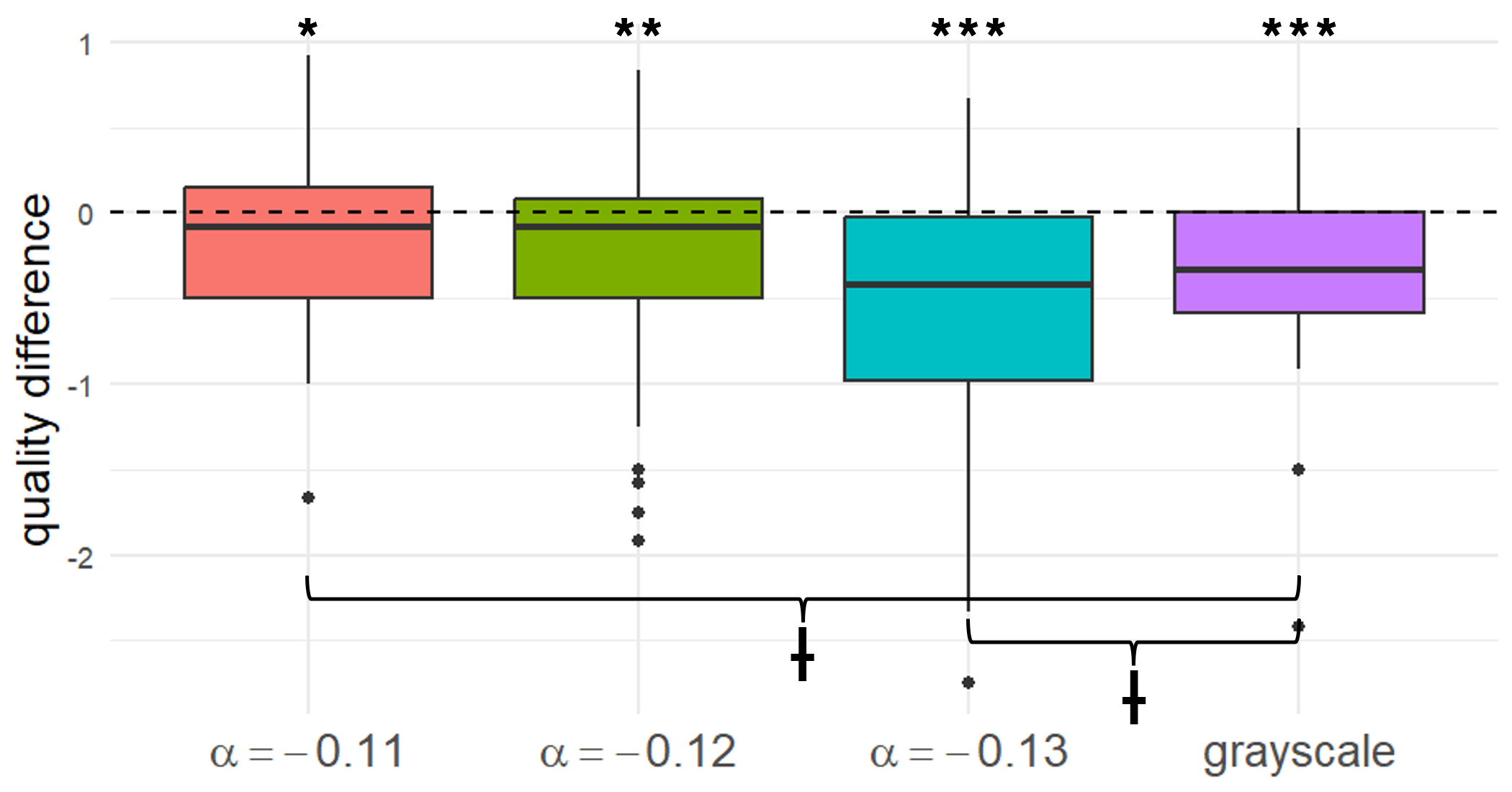}
    \caption{}
    \label{fig:diff-qua}
  \end{subfigure}
  \hfill
  \begin{subfigure}[t]{0.32\linewidth}
    \centering
    \includegraphics[width=\linewidth]{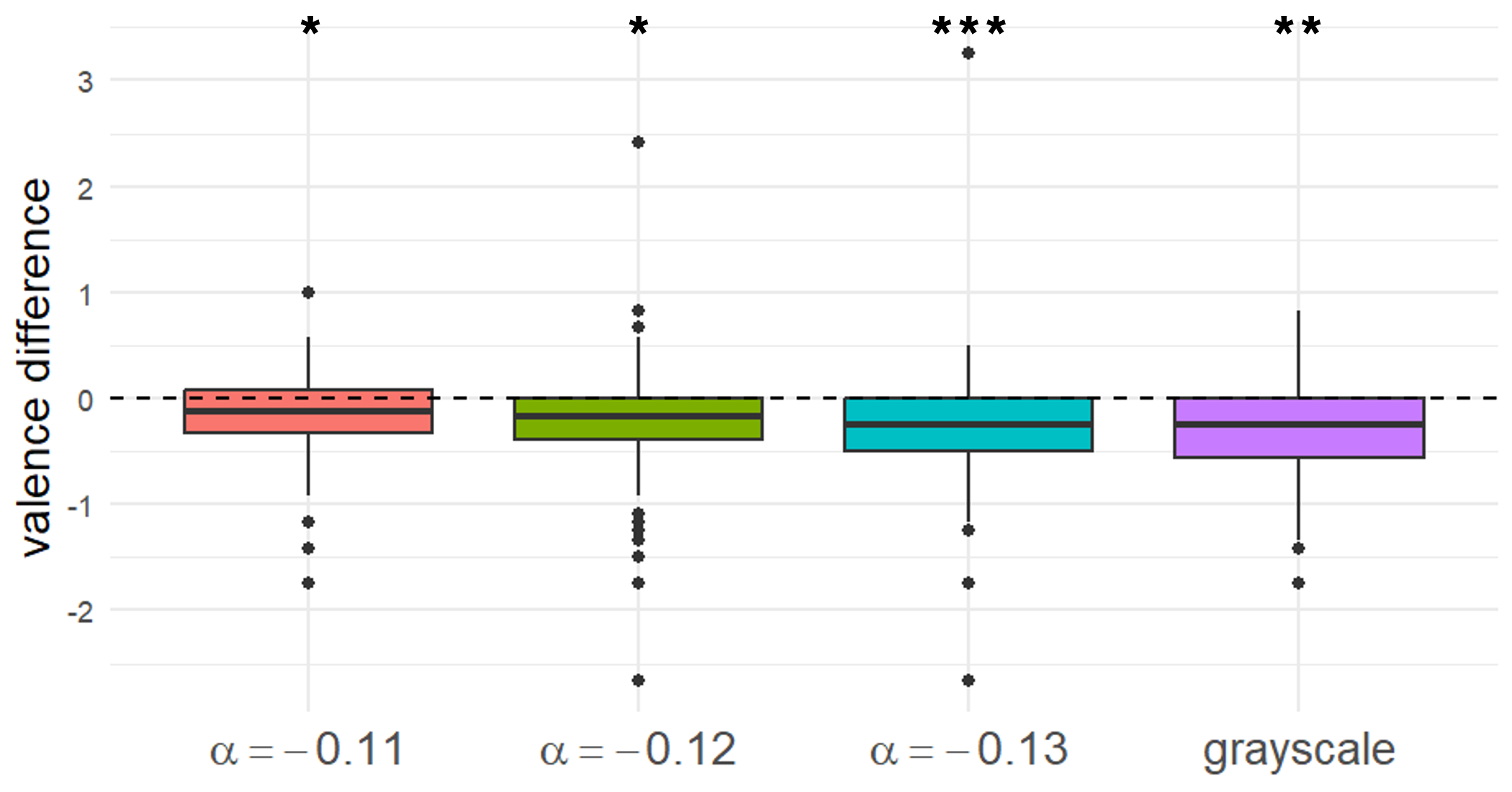}
    \caption{}
    \label{fig:diff-val}
  \end{subfigure}
  \caption{\textbf{Rating differences.} Per-participant rating differences between each condition and \emph{original} for (a) arousal, (b) perceived image quality, and (c) valence. Each box summarizes one difference score per participant; values below the dashed zero line indicate lower ratings than for the original images. Significance notation as in \cref{tab:msd}.}
  \label{fig:diffs}
\end{figure*}

\paragraph{Stimuli}
To ensure the external validity of our study, we used images from the Instagram Influencer Dataset \cite{kim2020multimodal}. 
It contains 10,180,500 Instagram posts from 33,935 influencers. 
Posts are classified into the following eight categories: beauty, family, fashion, fitness, food, interior, pet, and travel. 
To match preferences of a broader audience, we selected 17 images from the categories fitness and travel. 
All study stimuli were selected prior to inspecting model outputs, precluding cherry-picking.

Following the previously mentioned study design, we split the 17 images into 5 calibration images and a block of 12 stimulus images, which were presented in each condition. The image order was kept constant across conditions.
All 17 images (5 calibration and 12 stimulus images) across the five conditions are shown in \cref{fig:qualitative-results-1,fig:qualitative-results-2,fig:qualitative-results-3}.

\paragraph{Task and procedure}
Participants viewed each image for a minimum of five seconds, after which they could continue viewing or advance by pressing the space key. They then rated valence and arousal on a validated questionnaire~\cite{bradley1994measuring} and image quality on a Likert item adapted from~\cite{mould2012emotional}. A neutral stimulus followed each image--questionnaire pair to prevent emotional contamination. The task repeated for all 12 images in each condition.

Participants provided consent, completed a demographic questionnaire, and familiarized themselves with the setup in a trial run. After a two-minute relaxation video, they proceeded through the calibration images and the five condition blocks. Sessions lasted approximately 30 minutes and were compensated with £6.50.

\paragraph{Participants}
We conducted an a priori power analysis to establish the minimum sample size needed to test our hypotheses. 
To achieve 95\% statistical power at $\alpha = 0.05$ for a repeated-measures ANOVA with five conditions and a Cohen’s f of 0.2, we estimated a required sample size of $N = 48$. The chosen effect size was based on reported values in similar studies (0.2–1.5; \cite{de2010effects, bekhtereva2017bringing, Gebhardt2025Generative}).
To account for potential outliers or participants who might not complete the study faithfully, we recruited 57 participants (24 female, 33 male), ages 22--71 (M=38, SD=13.2) from the online crowd-sourcing platform Prolific. 
We excluded two participants whose data clearly indicated they did not perform the task faithfully (e.g., rating all images identically and failing attention checks), and one participant who experienced technical difficulties.
The data of the remaining 54 participants was used in the analysis.
Importantly, the significance of the results remains unchanged regardless of whether these outliers were included or excluded.

\paragraph{Ethics}
The study did not require institutional ethics board review under the applicable national and institutional regulations, as it falls outside the scope of national human-research legislation.
Participants were recruited via Prolific, gave informed consent prior to participation, were compensated for their time, and could withdraw at any point without penalty. No personally identifiable information was collected; all ratings were anonymous.

\paragraph{Statistical analysis}
We follow the analysis protocol of \cite{Gebhardt2025Generative}, whose design our study replicates. We conducted repeated-measures ANOVAs on each dependent variable; Mauchly's test indicated violations of sphericity for all measures (all $p<.001$), so we report Greenhouse–Geisser corrected degrees of freedom throughout.
Uncorrected and corrected ANOVAs never differed in significance.
For post-hoc comparisons, Shapiro–Wilk tests indicated non-normal paired differences for 20 of 26 comparisons; we therefore report Holm-corrected Wilcoxon signed-rank tests as the primary significance test, with Cohen's $d_z$ as the effect size for comparability with \cite{Gebhardt2025Generative}. For transparency, we additionally report the Holm-corrected paired $t$-test wherever the two tests diverged in significance.

\subsection{Results}
\label{apx:user-study:results}
In the following, we present the results of the study.
\cref{tab:msd} reports mean ratings per condition; \cref{fig:diffs} shows the corresponding per-participant differences from \cdefault{}.

We found an effect of the \ivfilter{} factor levels on arousal ratings \anova{3.07}{162.78}{2.78}{=}{.042}{.05}.
Post-hoc comparisons showed that \cgray{} caused significantly lower arousal ratings than \cdefault{} \pvald{=}{.002}{-.42}. 
The three adapted conditions also reduced arousal relative to \cdefault{}: \cone{} \pvald{=}{.021}{-.30}, \ctwo{} \pvald{=}{.021}{-.25}, \cthree{} \pvald{=}{.021}{-.28}.
For these three comparisons, the paired $t$-test did not reach significance (all \pvall{=}{.099}), consistent with the non-normal paired differences reported above.




The ANOVA also revealed a main effect of \ivfilter{} on perceived image quality \anova{2.78}{147.10}{18.15}{<}{.001}{.26}.
Pairwise comparisons showed that \cgray{} \pvald{<}{.001}{-.66}, \cthree{} \pvald{<}{.001}{-.80}, \ctwo{} \pvald{=}{.003}{-.47}, and \cone{} \pvald{=}{.014}{-.39} received significantly lower quality ratings than \cdefault{}.
The comparisons against \cgray{} revealed an asymmetric pattern across the adapted conditions: \cone{} was rated significantly higher in quality than \cgray{} \pvald{=}{.015}{.41}, whereas \cthree{} was rated significantly lower \pvald{=}{.021}{-.39}; \ctwo{} did not differ significantly from \cgray{} \pvald{=}{.179}{.15}.




In addition, we found an effect of the \ivfilter{} on valence ratings \anova{2.73}{144.62}{4.35}{=}{.007}{.08}.
Post-hoc pairwise comparisons showed that \cgray{} \pvald{=}{.004}{-.49}, \cthree{} \pvald{<}{.001}{-.42}, \ctwo{} \pvald{=}{.017}{-.31}, and \cone{} \pvald{=}{.017}{-.33} were associated with lower valence ratings than \cdefault{}. 
No differences were found in the pairwise valence comparisons with \cgray{}.



Following \cite{Gebhardt2025Generative}, we contrasted valence ratings against the scale midpoint (5), testing whether adaptation moves valence toward neutral. All conditions differed significantly from the midpoint (all \pvall{<}{.001}). 
The original images were rated above neutral ($M = 6.34$); adapted conditions shifted valence downward but remained above the midpoint.




\subsection{Discussion}
Our experiment tested whether the manipulations in each condition altered viewers' perceived arousal, image quality, and valence relative to \cdefault{}.

For arousal, the omnibus effect of \ivfilter{} was significant, and all conditions showed reductions relative to \cdefault{} of similar direction and magnitude ($d_z=-.25$ to $-.42$). All four comparisons were significant under Holm-corrected Wilcoxon tests. 
The parametric $t$-test reached significance only for \cgray{}, consistent with its reduced power under the non-normal paired differences observed for the adapted conditions.

We also observed a reduction in perceived image quality for all conditions relative to \cdefault{}, with the size of this cost varying considerably across $\alpha$. Notably, \cone{} was rated significantly higher in quality than \cgray{} (\cthree{} was rated significantly lower; \ctwo{} did not differ from \cgray{}). 
Combined with the arousal reduction observed for \cone{}, this indicates that appropriate parametrization of $\alpha$ can achieve an arousal-reducing effect at a lower perceptual-quality cost than grayscale conversion, with \cone{} representing the most favorable trade-off observed in this study.

For valence, all conditions received lower ratings than \cdefault{}. A potential explanation lies in the coupling between valence and arousal: affective image datasets commonly exhibit a V-shaped population-level relationship in which images of neutral valence tend to have the lowest arousal (\cref{apx:va-dataset}). Under this coupling, reducing arousal would be expected to pull valence toward the neutral point, consistent with the shifts we observe. 

While we have not examined whether these perceptual effects translate into behavioral outcomes, our results indicate a promising direction: minimally invasive image adaptations, at an appropriately chosen $\alpha$, can reduce arousal at a lower perceptual-quality cost than grayscale conversion, while preserving image identity. Testing whether these perceptual shifts translate into behavioral outcomes, such as reduced engagement, is an important next step.

\end{document}